\documentclass[conference]{IEEEtran}
\usepackage{times}

\usepackage[numbers]{natbib}
\usepackage{float}

\usepackage{times}
\usepackage{graphicx}
\usepackage{calc}
\usepackage[notransparent]{svg}
\usepackage{amssymb}
\usepackage{amsmath}
\usepackage{xcolor}
\definecolor{cvprblue}{rgb}{0.21,0.49,0.74}
\usepackage[bookmarks=true]{hyperref}
\hypersetup{
	colorlinks=true,
	linkcolor=orange,
	filecolor=magenta,      
	urlcolor=orange,
	citecolor=cvprblue,
}
\usepackage{cleveref}
\usepackage{booktabs} 
\usepackage{multirow} 
\usepackage{multicol} 
\usepackage[flushleft]{threeparttable}
\usepackage{colortbl} 
\usepackage{diagbox}  
\usepackage{siunitx}

\usepackage{xspace} 
\usepackage[subrefformat=parens,labelformat=parens,font=small]{subcaption} 
\usepackage{wrapfig}

\usepackage{algorithm}
\usepackage{listings}

\crefname{table}{Tab.}{Tabs.}
\crefname{figure}{Fig.}{Figs.}
\crefname{section}{Sec.}{Secs.}
\crefname{appendix}{Appendix.}{Sections.}

\def\ie{\emph{i.e.}\xspace}

\newcommand{\ours}{SpatialVLA\xspace}

\newcounter{boldpara}
\newcommand{\boldparagraph}[1]{
    \refstepcounter{boldpara}
    \vspace{0.1em}\noindent{\bf #1}
}

\definecolor{mblue}{HTML}{367dbd}
\colorlet{colorFst}{mblue!45}     
\colorlet{colorSnd}{mblue!25}     
\colorlet{colorTrd}{yellow!30}      
\colorlet{colorLow}{darkgray!30}    
\definecolor{R1}{HTML}{E97451}
\definecolor{R2}{HTML}{008080}
\definecolor{R3}{HTML}{0047AB}
\colorlet{cmt}{darkgray!80}         
\colorlet{supp}{darkgray!50}        

\DeclareMathOperator*{\argmin}{arg\,min}

\definecolor{AC}{HTML}{E97451}
\definecolor{R1}{HTML}{008080}
\definecolor{R2}{HTML}{0047AB}
\definecolor{R3}{HTML}{EE82EE}
\definecolor{R4}{HTML}{6A5ACD}
\definecolor{Q}{HTML}{1f618d}

\IEEEoverridecommandlockouts 

\begin{document}
\makeatletter
\let\@oldmaketitle\@maketitle
\renewcommand{\@maketitle}{\@oldmaketitle
  \begin{center}
  \captionsetup{type=figure}
  \setcounter{figure}{0}
  \includegraphics[trim=0.4ex 0 0 0, clip, width=1.0\textwidth]{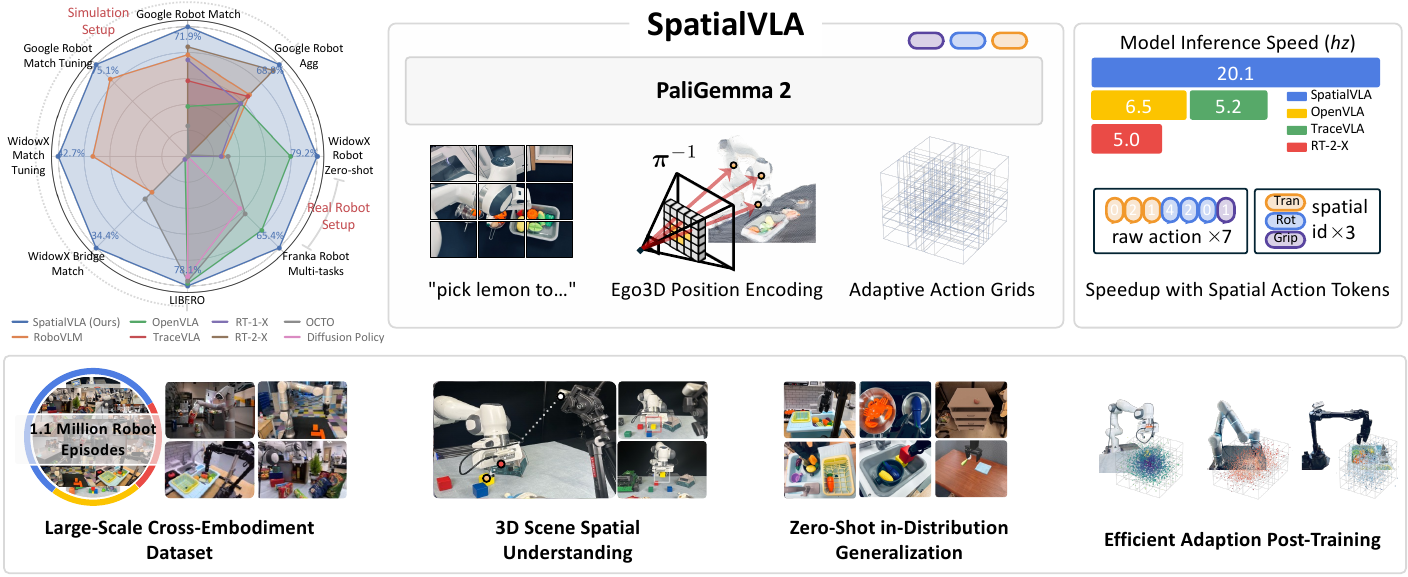}
    \caption{We present \ours, a spatial-enhanced vision-language-action model that is trained on 1.1 Million real robot episodes. The model is equipped with Ego3D Position Encoding and Adaptive Action Grids to explore spatial representations for generalist robot policies, achieving superior 3D scene spatial understanding, zero-shot in-distribution generalization, and efficient adaption to new robot setups. The model achieves state-of-the-art performance across a diverse range of evaluations and shows significantly faster inference speed with fewer tokens per action.} 
    \vspace{-1ex}
    \label{fig:teaser}
  \end{center}
}
\makeatother

\let\titleold\title
\renewcommand{\title}[1]{\titleold{#1}\newcommand{\thetitle}{#1}}
\def\maketitlesupplementary
{
    \newpage
    \begin{center}
        \Large
        \textbf{\thetitle}\\
        \vspace{0.5em}Supplementary Material \\
        \vspace{1.0em}
    \end{center}
}

\title{
\raisebox{-0.175cm}{\includegraphics[width=0.03\textwidth]{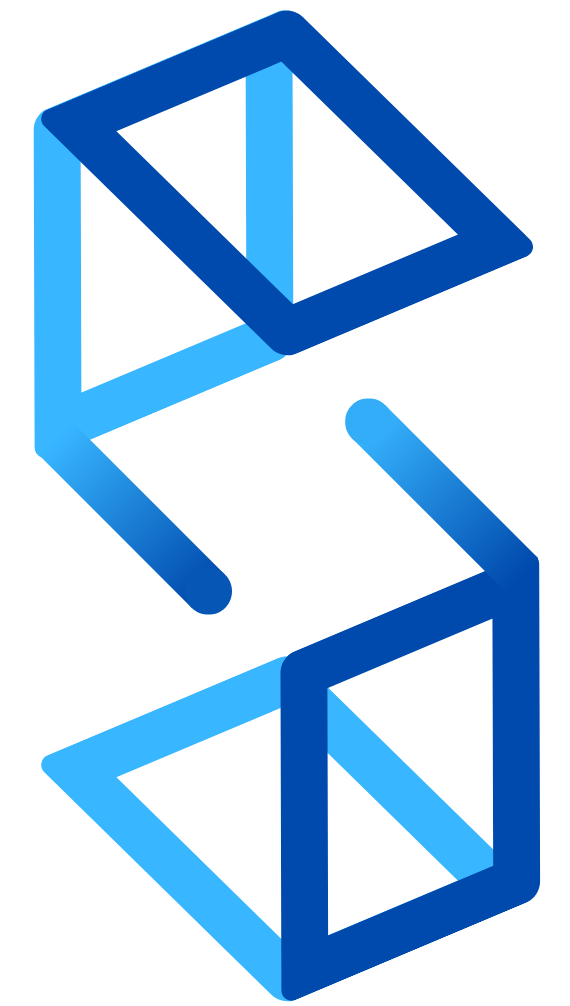}}
\textbf{\emph{SpatialVLA}}: Exploring Spatial Representations for Visual-Language-Action Model
}



\author{
  Delin Qu$^{\ast1,2}$, Haoming Song$^{\ast1,3}$, Qizhi Chen$^{\ast1,4}$, Dong Wang$^{\dagger1}$,
  Yuanqi Yao$^1$, Xinyi Ye$^1$, Yan Ding$^1$, \\ Zhigang Wang$^1$
  Jiayuan Gu$^5$, Bin Zhao$^{\dagger1,6}$, Xuelong Li$^{1,6}$ \\
  $^1$Shanghai AI Laboratory,
  $^2$Fudan University,
  $^3$Shanghai Jiao Tong University,
  $^4$Zhejiang University, \\
  $^5$ShanghaiTech University,
  $^6$Northwestern Polytechnical University \\
  \url{https://spatialvla.github.io}
  \thanks{
  $\ast$ Authors contributed equally: \texttt{dlqu22@m.fudan.edu.cn}. $\dagger$ Corresponding authors: \texttt{dongwang.dw93@gmail.com}.
  }
}

%

\maketitle

\begin{abstract}
    In this paper, we claim that spatial understanding is the keypoint in robot manipulation, and propose \ours to explore effective spatial representations for the robot foundation model. Specifically, we introduce \emph{Ego3D Position Encoding} to inject 3D information into the input observations of the visual-language-action model, and propose \emph{Adaptive Action Grids} to represent spatial robot movement actions with adaptive discretized action grids, facilitating learning generalizable and transferrable spatial action knowledge for cross-robot control. 
    \ours is first pre-trained on top of a vision-language model with 1.1 Million real-world robot episodes, to learn a generalist manipulation policy across multiple robot environments and tasks.
    After pre-training, \ours is directly applied to perform numerous tasks in a zero-shot manner. The superior results in both simulation and real-world robots demonstrate its advantage of inferring complex robot motion trajectories and its strong in-domain multi-task generalization ability. We further show the proposed \emph{Adaptive Action Grids} offer a new and effective way to fine-tune the pre-trained \ours model for new simulation and real-world setups, where the pre-learned action grids are re-discretized to capture robot-specific spatial action movements of new setups. The superior results from extensive evaluations demonstrate the exceptional in-distribution generalization and out-of-distribution adaptation capability, highlighting the crucial benefit of the proposed spatial-aware representations for generalist robot policy learning.
    All the details and codes are open-sourced.
\end{abstract}    
\IEEEpeerreviewmaketitle
\section{Introduction}
\label{sec:intro}
Generalist robot policies that are capable of interacting with the physical environment, adapting to various embodiments, and performing complex tasks have been a long-standing pursuit in robotics~\citep{brohan2022rt, bharadhwaj2024roboagent, doshi2024scaling, cheang2024gr, wang2024scaling}. Recent advances in Vision-Language-Action (VLA) models~\citep{brohan2023rt, kim2024openvla, black2024pi_0, li2024towards} show a promising paradigm in building such generalist policy by fine-tuning the pre-trained Vision-Language Models (VLMs)~\citep{alayrac2022flamingo,radford2021learning,peng2023kosmos,liu2024visual} on diverse robot data~\cite{o2024open, khazatsky2024droid, fang2023rh20t}. The key to the success of this paradigm lies in adapting the generalization power of VLMs to numerous robot manipulation tasks, as well as specific architectural designs that synergize the VLM backbone and robot action output head. Nonetheless, existing VLA models are primarily confined to 2D observation inputs and lack precise perception and comprehension of the 3D physical world — where humans instinctively construct rich, structured mental representations of space, effortlessly aligning objects within a canonical, intuitive, and even personally tailored workspace for manipulation~\citep{gallistel1990organization,logie2014visuo,piaget2013child,tolman1948cognitive,yang2024thinking}.
Therefore, an essential question for the field now is \emph{\textbf{how to effectively equip the VLA models with a profound spatial understanding of the 3D physical world?}}

However, developing such generalist robot policies with 3D spatial intelligence encounters two primary challenges in the aspects of robot observation and action. Firstly, the observations from different robot embodiments are not 3D-aligned, because the camera sensors of different robots are various and mounted at different places (\emph{e.g.} wrist and/or third-person), resulting in non-calibrated 3D observation spaces. Secondly, different robots have different action movement characteristics to accomplish diverse tasks, due to different degrees of freedom, motion controllers, workspace configurations, and task complexity, leading to significant difficulty in learning generalizable spatial actions.
Despite some attempts in generalist policy learning across heterogeneous robots~\citep{team2024octo,o2024open,kim2024openvla,wang2024scaling}, advancement in 3D spatial understanding abilities of generalist policy has significantly lagged behind. This is largely attributed to the heterogeneity in robot observation and action information. The solutions to the above challenges require spatial-aligned robot observation and action representations for cross-embodiment control and adaptation in the universal 3D physical world. 

In this work, as illustrated in~\cref{fig:teaser}, we propose a generalist robot policy \ours, which equips the VLA model with 3D spatial intelligence by exploring aligned spatial representations of robot observation and action signals. \ours perceives 3D world through \emph{\textbf{Egocentric 3D (Ego3D) Position Encoding}} to integrate 3D spatial context with semantic features. This position encoding is derived in the egocentric camera frame that eliminates the need for specific robot-camera calibration, which is universally applicable to various robot embodiments. As for robot actions, \ours unifies the action space of various robots via \emph{\textbf{Adaptive Action Grids}}, which discretizes the continuous robot actions into adaptive spatial grids according to statistical action distributions on the whole robot episodes and learns spatial action tokens on these grids to align cross-robot actions with the 3D spatial structure of the physical world. Crucially, after pre-training, the learned spatial action grids demonstrate a superior capability in adapting to new robot environments via adaptively grid re-discretization, providing a flexible and effective approach to robot-specific post-training. We find that the proposed model \ours bridges observation inputs and action outputs in a universal robot-agnostic manner, which explores powerful 3D spatial-aware representations to enhance the VLA model.


We extensively evaluate and ablate \ours on diverse robot manipulation tasks and different robot embodiments in both simulation and real-world, including 24 real-robot tasks and 3 simulation environments. To broadly test \ours as a generalist robot policy, we examine the model's abilities in zero-shot in-distribution robot control and new robot setup adaption abilities with instruction following, 3D scene structure understanding, and fine-tuning to new robot environments. The evaluation setups include view/texture/lighting change, unseen objects, unseen robot environment, and challenging spatial layout changes in robot setups and environments, demonstrating remarkable generalizability and transferability of \ours with spatial-aware representations. In summary, the contributions of this work consist of a novel generalist robot policy that explores spatial representations for robot foundation models, sophisticated designs on Ego3D Position Encoding and Adaptive Action Grids for effective 3D-awareness injection, and superior evaluation results across various robot setups and tasks.

    


\section{Related Work}
\label{sec:relate_work}
\begin{figure*}[ht]
    \vspace{-4ex}
    \begin{center}
        \includegraphics[trim=0.4ex 0 0 0, clip, width=1\linewidth]{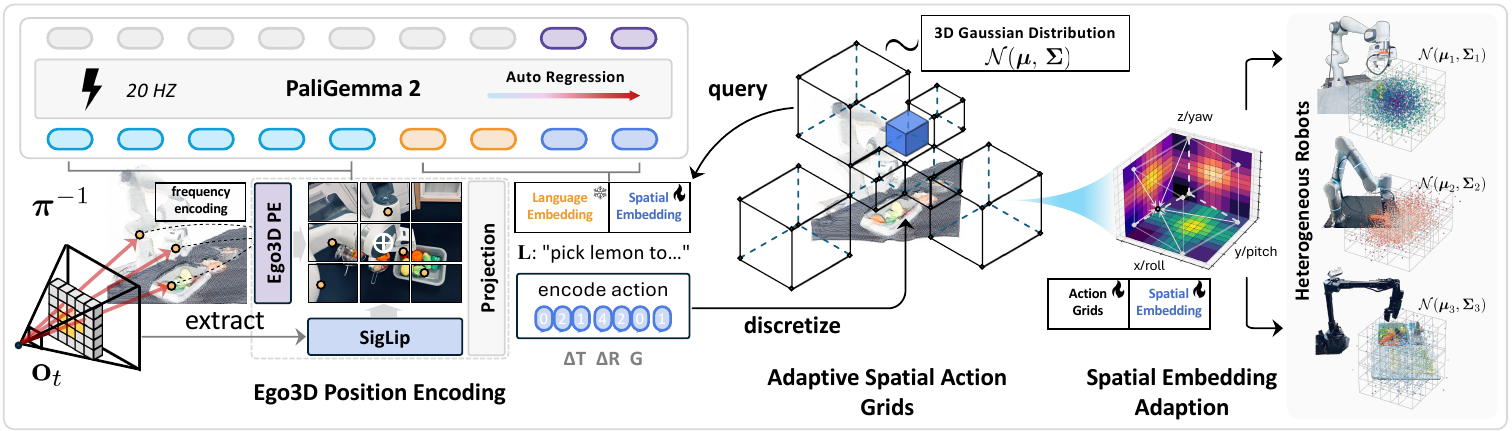}
    \end{center}
    \vspace{-1.5ex}
    \caption{
        \textbf{Overview of \ours}. Given an image observation $\mathbf{o}_t$ and a task instruction $\mathbf{L}$, the model processes the image using Ego3D Position Encoding and auto-regressively predicts spatial action tokens, which are then de-tokenized to generate continuous actions $\mathbf{A}_t$ for robot control. The model comprises three key components: (1) SigLIP vision encoder extracts 2D semantic features, which are then infused with 3D spatial context via Ego3D Position Encoding; (2) continuous 7D actions ${\Delta\mathbf{T}, \Delta\mathbf{R}, \mathbf{G} }$ are translated to 3 spatial action tokens by querying Adaptive Action Grids and auto-regressively predicted and de-tokenized for robot control; (3) in post-training, action grids and spatial embeddings are adapted from new Gaussian distributions to facilitate effective transfer to new robot setups.
    }
    \label{fig:pipeline}
    \vspace{-2.5ex}
\end{figure*}

\boldparagraph{Generalist Robot Polices.}
Recent advances in robotics have witnessed a trend towards developing multi-task "generalist" robot policies to perform diverse tasks, rather than one specific task. Some early works~\citep{parisotto2015actor, rusu2015policy, haldar2023polytask, brohan2022rt, shridhar2023perceiver, zhu2023learning, haldar2024baku} achieve great success in learning a language-conditioned visual multi-task policy on a single embodiment with pre-trained visual/text encoder, thereby lacking the ability to adapt new robot embodiment. More recent efforts~\citep{team2024octo, liu2024rdt, wang2024scaling} explore to use large-scale, cross-embodiment robot datasets~\citep{o2024open} for generalist polices pre-training, supporting effective fine-tuning to new robot setups. Notably, Octo~\citep{team2024octo} proposes a flexible transformer-based architecture to unify different configurations in Open X-Embodiment (OXE) dataset~\citep{o2024open}, and the trained policy can solve a variety of in-domain tasks in zero-shot and achieves strong performance in the new embodiment after fine-tuning. With the same cross-embodiment robot datasets, RDT~\citep{liu2024rdt} pre-trains a 1.2B-parameter diffusion-based generalist model and fine-tunes it for complex bimanual manipulation. Moreover, HPT~\citep{wang2024scaling} proposes a modular architecture to align data across heterogeneous embodiments into a shared representation via embodiment-specific stem module, embracing the heterogeneity in data through pre-training.

\boldparagraph{Vision-Language-Action Models.}
Recently, several studies~\citep{livision,brohan2023rt,kim2024openvla,li2024cogact,zheng2024tracevla,li2024towards,zheng2025universal,pertsch2025fast} propose to build generalist robot policies by extending pre-trained VLMs with ability to robot action generation. As a pioneer, RT-2~\citep{brohan2023rt} fine-tune VLM PaLI-X~\citep{chen2023pali} on both large-scale vision-language data and robot demonstration data via autoregressive next token prediction, where robot actions are discretized into 256 bins and represented as separate tokens analogous to text tokens. OpenVLA~\citep{kim2024openvla} adopts a similar action discretization approach and fine-tune Prismatic VLM~\citep{karamcheti2024prismatic} only on the OXE dataset~\citep{o2024open}, which consists of robot data from 22 different robot embodiments across 21 institutions. CogACT~\citep{li2024cogact} and TraceVLA~\citep{zheng2024tracevla} continue to fine-tune the trained OpenVLA model with the new attached diffusion action module and visual trace prompting separately. Moreover, $\pi_0$~\citep{black2024pi_0} adapts PaliGemma VLM to robot control by adding a separate action expert module that produces continuous actions via flow matching, and the model can then be prompted for zero-shot control or fine-tuned on high-quality data to enable complex dexterous manipulation tasks. Notably, while these models benefit from VLMs’ capabilities and show some zero-shot capabilities, a sophisticated fine-tuning step with new data is essential and required for complex tasks or new robot setups.

\boldparagraph{3D Foundation Models for Robotics.} 
Some researches~\citep{zhu2024llava,chen2024ll3da,fu2024scene,hong20233d,qu2024livescene,huang2023embodied,zhen20243d} have focused on extending the generalist ability of LLMs and VLMs from language-vision towards the 3D world. 3D-LLM~\citep{hong20233d} integrates a 3D feature extractor with 2D VLMs backbone and train 3D-LLMs on a wide variety of tasks, including dense captioning, 3D question answering, task decomposition, 3D grounding, 3D-assisted dialog, navigation, and so on. 
LLaVA-3D~\citep{chen2024ll3da} extends the 2D LLaVA’s capabilities with the proposed 3D patches to bridge 2D features within a 3D space for 3D spatial understanding. Similarly, LEO~\citep{huang2023embodied} trains an embodied multi-modal generalist agent that can take egocentric 2D images, 3D point clouds, and texts as task input and handle comprehensive tasks within the 3D environment. Moreover, 3D-VLA~\citep{zhen20243d} builds a generative world model on top of 3D-based LLM to perform 3D reasoning and localization, multimodal goal generation, and embodied action planning. LEO and 3D-VLA are closely related to our work, but their attention is on 3D world understanding and prediction, ignoring the 3D spatial characteristics in the robot action space.

\section{Methodology}
\label{sec:method}
In this section, we describe \ours model and its training framework in detail. Our model with the proposed Ego3D position encoding and adaptive action grids to capture and learn 3D spatial knowledge for generalizable robot control, which we describe in~\cref{sec:model}. 
Next, we detail the training procedure of \ours that consists of a pre-training stage and a post-training stage in~\cref{sec:training}. The pre-training aims to learn generalizable knowledge with large-scale cross-robot data and the goal of post-training is to adapt pre-trained model to specific downstream robot embodiments and tasks.

\subsection{The \ours Model Architecture}
\label{sec:model}
As illustrated in~\cref{fig:pipeline}, \ours is developed based on a vision-language model to inherit the general world knowledge. Formally, \ours takes image observations $\mathbf{o}_t=\{\mathbf{I}_t^1,..,\mathbf{I}_t^n\}$ and a natural language task instruction $\mathbf{L}$ as inputs, and then learns a mapping function $\tau(\cdot)$ to generate a sequence of robot actions $\mathbf{A}_t=[\mathbf{a}_t, \mathbf{a}_{t+1}, ..., \mathbf{a}_{t+H-1}]$, \ie, $\mathbf{A}_t=\mathcal{F}(\mathbf{o}_t, L)$. To empower \ours with 3D spatial intelligence, we augment the VLM backbone with robotics-specific 3D-aware inputs and outputs, namely,  \emph{Ego3D Position Encoding} and \emph{Adaptive Action Grids}. The ego3D position encoding representation $\mathbf{O_{3d}}$ aims to capture 3D scene structure via integrating 3D spatial information with 2D semantic features. The adaptive action grids are designed to represent the continuous distribution of robot actions $\mathbf{a}$ with a set of discrete spatial action tokens $\mathfrak{a}=\{\mathfrak{a}^1,...,\mathfrak{a}^{V}\}$. During training, \ours model is trained to take the ego3D position encoding representation $\mathbf{O_{3d}}$ and natural language task instruction $\mathbf{L}$ as inputs, and autoregressively generate spatial action tokens $\tilde{\mathfrak{a}}_t$ using the cross-entropy objective $\mathcal{L}$,
\begin{equation}
\label{eq:loss}
    \mathfrak{L}(\theta) = \mathbb{E}_{p(\mathbf{A}_t|\mathbf{o}_t)}\mathcal{L}(\mathfrak{a}_t, \tilde{\mathfrak{a}}_t)),
\end{equation}
where the predicted action tokens $\tilde{\mathfrak{a}}_t = \tau(\mathbf{O_{3d}},\mathbf{L})$ is the de-tokenized into continuous action signals $\mathbf{a}_t$ for robot control. More details of the model architecture and action encoding can be found in~\cref{supp:model_arch}.

\noindent\textbf{Ego3D Position Encoding.}
The proposed Ego3D position encoding integrates depth information from the camera frame and image pixels to construct an egocentric 3D coordinate system, which eliminates the need for robot-camera extrinsic calibration and is agnostic to specific robot setups. Specifically, we use ZoeDepth~\citep{bhat2023zoedepth} to estimate depth map $D$ and obtain pixel's 3D position $\mathbf{p}=\{x, y, z\}$ in the egocentric 3D coordinate system via back-projection $\pi^{-1}$ with camera intrinsic parameters. Then, as illustrated in~\cref{fig:pipeline}, we first employ SigLIP~\citep{zhai2023sigmoid} visual encoder to extract 2D semantic visual features $\mathbf{X} \in \mathbb{R}^{d\times h\times w}$ to inherit the alignment between vision and language, and calculate their corresponding 3D positions $\mathbf{P} \in \mathbb{R}^{3\times h\times w}$ in the egocentric 3D coordinate system. The egocentric 3D positions $\mathbf{P}$ are then encoded into 3D position embeddings $\mathbf{P}^{'}\in \mathbb{R}^{d\times h\times w}$ through a sinusoidal function $\boldsymbol{\gamma}(\cdot)$ following by a learnable MLP. The egocentric 3D spatial representations $\mathbf{O_{3d}} \in \mathbb{R}^{d\times h\times w}$ are obtained by adding 3D position embedding $\mathbf{P}^{'}$ and 2D path visual tokens $\mathbf{X}$, depicted as follows,
\begin{equation}
\label{eq:ego}
    \mathbf{O_{3d}} = \mathbf{X} + \mathbf{P}^{'} = \mathbf{X} + \text{MLP}(\boldsymbol{\gamma}(\mathbf{P})).
\end{equation}

\begin{figure}[t]
    \begin{center}
        \includegraphics[width=1\linewidth]{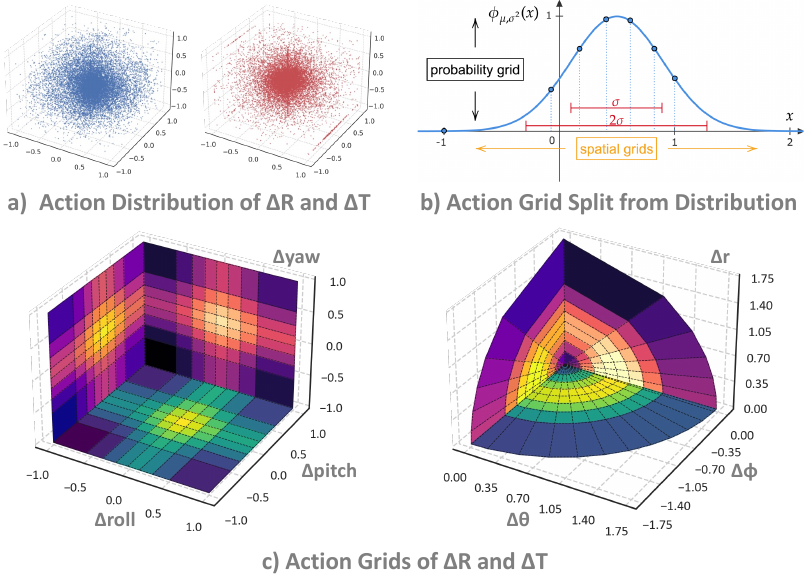}
    \end{center}
    \caption{Illustration of adaptive action grids. (a) Statistics of translation and rotation action movements on the whole pre-training mixture, (b) grids are split on each action variable according to the probability density function of fitted Gaussian distribution, and (c) the obtained adaptive action grids in translation and rotation action spaces.}
    \label{fig:bin}
\end{figure}

\noindent\textbf{Adaptive Action Grids.}
In order to auto-regressively generate continuous robot actions with pre-trained VLM backbone, we design Adaptive Action Grids to translate continuous robot actions to discrete grids that are represented as tokenized classes for prediction. 
Specifically, for a single-arm robot, its actions consist of seven dimensions for movement $\mathbf{a}=\{\text{x, y, z, roll, pitch, yaw, grip}\}$, and are split into three parts as follows,
\begin{equation}
\label{eq:action}
    \mathbf{a} = \{\mathbf{a}_\text{trans}, \; \mathbf{a}_\text{rot}, \; \mathbf{a}_\text{grip}\},
\end{equation}
where $\mathbf{a}_\text{trans}=\{\text{x, y, z}\}$ represents translation movements $\Delta \mathbf{T}$, $\mathbf{a}_\text{rot}=\{\text{roll, pitch, yaw}\}$ denotes rotation movements $\Delta \mathbf{R}$, and $\mathbf{a}_\text{grip}=\{\text{grip}\}$ consists of two discrete tokens that represent opening and closing gripper actions. Moreover, we transform the translation movements $\text{(x, y, z)}$ into polar coordinates ($\phi, \theta, r$) to disentangle movement direction ($\phi, \theta$) and distance $r$. 


As illustrated in~\cref{fig:bin}, for tokenizing continuous translation and rotation movements, we first normalize each action variable into $[-1, 1]$ for each robot environment and statistic the translation and rotation movements $\Delta \mathbf{R} = \{\text{roll, pitch, yaw}\}, \; \Delta \mathbf{T} = \{\text{$\phi, \theta, r$}\}$ on the whole dataset mixture (see~\cref{supp:stat}), following with a parameterized Gaussian distribution fitting $\mathcal{N}(\mu^{a}, \Sigma^{a})$. Then, the continuous actions are split into $\text{M}$ intervals $\mathbf{G}_{i=1,..,\text{M}}=\{[a_1=-1,a_2),...,[a_\text{M-1}, a_\text{M}=1]\}$ with equal probability $1/\text{M}$ on each normalized action variable, \ie, 
\begin{equation}
\label{eq:bin}
    a_2,...,a_\text{M} = \argmin_{a_2,..., a_\text{M}} \int_{a_i}^{a_{i+1}} f(x)dx - 1/\text{M}, \; i=1,...,\text{M}
\end{equation}
where $f(x)$ is the probability density function of Gaussian distribution $\mathcal{N}(\mu^{a}, \Sigma^{a})$. Note that we split more grids on $\{\text{$\phi, \theta$}\}$ to capture fine-grained movement direction other than movement distance $r$.
Suppose $\text{M}_\phi, \; \text{M}_\theta, \; \text{M}_r$ are the numbers of the discrete bins on variable ($\phi, \theta, r$), then the translation space consists of $\text{M}_\text{trans}=\text{M}_\phi \cdot \text{M}_\theta \cdot \text{M}_r$ discrete spatial grids $\mathfrak{a}_\text{trans}=\{\mathfrak{a}^1,...,\mathfrak{a}^{\text{M}_\text{trans}}\}$. Similarly, there are $\text{M}_\text{rot}=\text{M}_\text{roll} \cdot \text{M}_\text{pitch} \cdot \text{M}_\text{yaw}$ 3D discrete grids $\mathfrak{a}_\text{rot}=\{\mathfrak{a}^1,...,\mathfrak{a}^{\text{M}_\text{rot}}\}$ in rotation 3D spatial space. Then, the associated learnable spatial action token embeddings are defined as follows,
\begin{equation}
\label{eq:token}
    \mathbf{E}_{\mathfrak{a}} = \{\mathbf{E}_\text{trans}, \; \mathbf{E}_\text{rot}, \; \mathbf{E}_\text{grip}\},
\end{equation}
where $\mathbf{E}_\text{trans} \in \mathbb{R}^{d\times \text{M}_\text{trans}}$, \; $\mathbf{E}_\text{rot} \in \mathbb{R}^{d\times \text{M}_\text{rot}}$, \; $\mathbf{E}_\text{grip} \in \mathbb{R}^{d\times 2}$ denote the translation, rotation, and gripper actions, and the total number of action tokens is $V=\text{M}_\text{trans}+\text{M}_\text{rot}+2$. After training, these learned spatial action tokens capture general robot action knowledge and show a surprising ability in new robot embodiment adaption, as discussed in~\cref{sec:training}.
Moreover, it is worth noting that the model only needs to generate 3 tokens for one-step robot actions rather than 7 tokens as in RT-1~\cite{brohan2022rt}, RT-2~\cite{brohan2023rt} and OpenVLA~\cite{kim2024openvla}, achieving in fast model inference speed.

\subsection{The Pre-training and Post-training Scheme}
\label{sec:training}
To obtain a generalist robot policy model, the training procedure of \ours consists of pre-training stage and post-training stage. Pre-training stage aims to learn generalizable knowledge across diverse tasks and robots from a large-scale dataset mixture, while the post-training stage adapts the pre-trained model into new robot embodiments or new tasks. In the following, we discuss the dataset mixture and key designs for implementing this two-stage training procedure.

\noindent\textbf{Pre-training Procedure.}
We train \ours from Paligemma2 backbone~\citep{steiner2024paligemma} on a cross-robot dataset mixture with 1.1 Million real robot demonstrations $\{\zeta_1,...,\zeta_n\}$, covering a diverse range of robot embodiments, scenes, and tasks. This pre-training dataset mixture consists of a subset of OXE~\citep{o2024open} and the RH20T dataset~\citep{fang2023rh20t} and we modify the mixture weights from OpenVLA~\citep{kim2024openvla} according to the real-word testing performance of individual dataset, which are exhibited in~\cref{supp:data_mixture}. 
At the beginning of pre-training, the embeddings $\mathbf{E}_\mathfrak{a}$ of spatial action tokens and parameters of MLP in egocentric 3D spatial representation are randomly initialized, and then they are optimized during training, as well as the parameters of vision encoder and LLM backbone. At each training step, a batch of data pairs is extracted at random timesteps $t_1,...,t_{B}$ from shuffled demonstrations $\{\zeta_i,...,\zeta_j\}$, \ie, a batch of tuple $[(\mathbf{o}_{t_1}, \mathbf{A}_{t_1}, \mathbf{L}_i),...,(\mathbf{o}_{t_B}, \mathbf{A}_{t_B}, \mathbf{L}_j)]$, and then \ours is trained with a standard auto-regressive next-token prediction objective in~\cref{eq:loss}. Importantly, the embeddings of text tokens $\mathbf{E}_\text{text}$ are frozen to maintain the general world knowledge in pre-trained VLM, and the experimental results show this frozen operation is beneficial for the instruction following ability. Moreover, as discussed in OpenVLA~\citep{kim2024openvla}, DROID dataset~\citep{khazatsky2024droid} are removed from the data mixture for the final third of pre-training to improve the quality of the pre-trained \ours model.

\noindent\textbf{Post-training Designs.}
In the post-training stage, we fine-tune our model with robot-specific demonstrations to adapt it to new tasks and robot setups. Prior works have mainly focused on fine-tuning pre-trained VLA models using full-parameter or LoRA fine-tuning, with little attention to effective techniques for the post-training stage. In this paper, we investigate the potentials of the proposed spatial action tokenizer for quick adaption to new robot setups, namely \textbf{\textit{Spatial Embedding Adaption}}, providing a new and effective way for useful post-training. In detail, we fit a new Gaussian distribution $\mathcal{N}(\mu_\text{new}, \Sigma_\text{new})$ for each action variable on post-training datasets and create discrete spatial action grids $\mathbf{G}_\text{new}$ in translation and rotation movement to construct action grids $\mathbf{G}_\text{new}$ and tokens $\mathfrak{a}_\text{new}$, where the embeddings of new spatial action tokens $\mathbf{E}_{\mathfrak{a}_\text{new}}$ are initialized by trilinear interpolation with pre-trained action tokens $\mathbf{E}_\mathfrak{a}$. These action token embeddings $\mathbf{E}_{\mathfrak{a}_\text{new}}$ and model parameters are then optimized with the next-token prediction objective.  

Formally, for new spatial action grids $\mathbf{G}_\text{new}$, suppose $i$-th 3D grid $\mathbf{G}_\text{new}^i$ in translation space $\mathbf{a}_\text{trans}^\text{new}$ with centroid ($\phi_\text{new}^i, \; \theta_\text{new}^i, \; r_\text{new}^i$) and its adjacent 3D grids from the pre-trained action grids are $\mathbf{G}^\text{adj}=\{\mathbf{G}^1,..., \mathbf{G}^K\}$. The embedding of new $i$-th action token $\mathbf{e}_{\mathfrak{a}_\text{new}}^i$ are initialized by trilinear interpolation with $\mathbf{G}^\text{adj}$, as follows,
\begin{equation}
\label{eq:post-train}
    \mathbf{e}_{\mathfrak{a}_\text{new}}^i =  \sum_{j=1}^K w_j\mathbf{e}^j,
\end{equation}
where $\mathbf{e}^j_\mathfrak{a} \in \mathbb{R}^{d}$ are the embeddings of the pre-trained action grids, $w_j$ is the weights calculated by the normalized distances between centroid ($\phi_\text{new}^i, \; \theta_\text{new}^i, \; r_\text{new}^i$) and adjacent centroids. Note that the new action tokens of rotation $\mathbf{E}_{\mathfrak{a}_\text{rot}}^\text{new}$ are initialized in the same way. With this embedding initialization, the new action tokenizer is capable of effectively transferring pre-trained spatial action knowledge to new robot setups. 
\section{Experiment}
\label{sec:experiment}
The goal of our experimental evaluations is to test \ours's ability to serve as a generalist robot control policy out of the box, as well as be a good initialization for fine-tuning to new robot tasks. Our extensive experiments consist of zero-shot evaluations and adaption to downstream tasks in both simulation and real-world. \ours is compared to previous state-of-the-art robot foundation models and alternative designs in spatial representations. Concretely, experiments seek to answer the following research questions:
\begin{enumerate}
    \item How well does \ours directly perform on a variety of in-distribution tasks after pre-training on large-scale robotic data mixture?    

    \item Can \ours be effectively fine-tuned on new robot setup and task?
    
    \item How well does \ours perform in scenarios that require spatial understanding?


    \item To what extent do Egocentric 3D Spatial Representations and Adaptive Spatial Action Grids improve the performance of \ours?
\end{enumerate}

\begin{figure*}[t]
    \begin{center}
        \includegraphics[width=\linewidth]{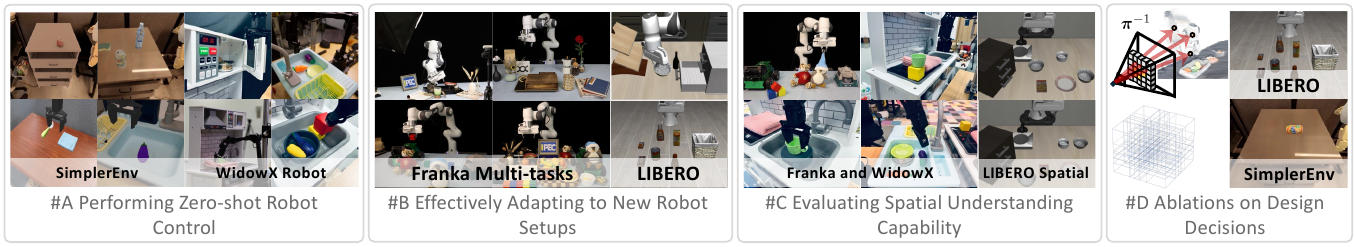}
    \end{center}
    \vspace{-1.5ex}
    \caption{\textbf{Experiment Setup.} We evaluate \ours across 7 robot learning scenarios, 16 real-robot tasks, and 48 simulation setups, focusing on three key aspects: zero-shot control, adaptability to new setups, and spatial understanding. We also conduct a thorough ablation study on a mixed Fractal and Bridge dataset to verify our design decisions.}
    \label{fig:experiment_overview}
\end{figure*}

To answer these questions, as shown in~\cref{fig:experiment_overview}, we evaluate \ours's capabilities across a representative spectrum of 7 different robot learning scenarios with 24 real-robot tasks and 3 simulation environments. Firstly, we evaluate \ours in both SimplerEnv~\cite{li24simpler} simulation and the real-world WidowX robot platform (BridgeV2~\citep{walke2023bridgedata}~\citep{walke2023bridgedata} setups), testing its out-of-the-box control capabilities on different robots with setups matching the pre-training dataset. Second, we assess the fine-tuning efficacy of our method in both simulation and real-world settings, including LIBERO~\citep{liu2023libero} and new Franka robot setups, to adapt to new robot environments and tasks. Then, we design 4 special tasks that require precise spatial understanding in 2 different real-world robot environments to test the effectiveness of spatial representations of \ours. Finally, we conduct comprehensive ablation studies on a mixture of Fractal~\citep{brohan2022rt} and BridgeV2~\citep{walke2023bridgedata} datasets to verify our design decisions in \ours. For more details on evaluation setups, see~\cref{supp:eval_setup}.

\noindent\textbf{Implementation Details.}
The \ours model is pre-trained with 1.1 Million real-robot demonstrations from the OXE ~\citep{o2024open} and RH20T dataset~\citep{fang2023rh20t} on a cluster of 64 A100 GPUs for 10 days, using a batch size of 2048. For input robot observation, the \ours policy is only conditioned on one third-person camera and takes one image for constructing egocentric 3D spatial representations. For output robot actions, the \ours policy predicts a chunk of $T=4$ future actions (12 spatial action tokens from total $V=8194$ tokens) and executes the ensemble actions before predicting the next chunk. During inference, \ours requires 8.5GB of GPU memory and runs at approximately 20Hz on one NVIDIA RTX 4090 GPU to run evaluations in both simulation and real-world. For more details about model training and deployment, please refer to the~\cref{supp:model_pretrain_deploy}.

\begin{table*}
    [t]
    \centering
    \caption{\textbf{SimplerEnv evaluation across different policies on Google
            Robot tasks}. The zero-shot and fine-tuning results denote performance of OXE
        dataset~\citep{o2024open} pre-trained models and Fractal dataset~\citep{brohan2022rt}
        fine-tuned models, respectively.}
    \label{tab:simplerenv_google_robot} \resizebox{2\columnwidth}{!}{
        \begin{threeparttable}
            \begin{tabular}{l|cccccccccc}
                \toprule \multirow{2}{*}{Model}                    & \multicolumn{4}{c}{\textbf{Visual Matching}} & \multicolumn{4}{c}{\textbf{Variant Aggregation}}                                                                                                                                                                                                  \\
                \cline{2-5} \cline{7-10}                           & Pick Coke Can                                & Move Near                                        & \begin{tabular}[c]{@{}l@{}}Open/Close Drawer\end{tabular} & \#Average       &  & Pick Coke Can  & Move Near      & \begin{tabular}[c]{@{}l@{}}Open/Close Drawer\end{tabular} & \#Average       \\
                \cmidrule{1-10} RT-1~\citep{brohan2022rt} (Begin)  & 2.7\%                                        & 5.0\%                                            & 13.9\%                                                    & 6.8\%           &  & 2.2\%          & 4.0\%          & 6.9\%                                                     & 4.2\%           \\
                RT-1~\citep{brohan2022rt} ($15\%$)                 & 71.0\%                                       & 35.4\%                                           & 56.5\%                                                    & 60.2\%          &  & 81.3\%         & 44.6\%         & 26.7\%                                                    & 56.2\%          \\
                RT-1~\citep{brohan2022rt} (Converged)              & 85.7\%                                       & 44.2\%                                           & 73.0\%                                                    & 74.6\%          &  & 89.8\%         & 50.0\%         & 32.3\%                                                    & 63.3\%          \\
                \cmidrule{1-10} HPT~\citep{wang2024scaling}        & 56.0\%                                       & 60.0\%                                           & 24.0\%                                                    & 46.0\%          &  & -------------- & -------------- & --------------                                            & --------------  \\
                TraceVLA~\cite{zheng2024tracevla}                  & 28.0\%                                       & 53.7\%                                           & 57.0\%                                                    & 42.0\%          &  & 60.0\%         & 56.4\%         & 31.0\%                                                    & 45.0\%          \\
                RT-1-X~\cite{o2024open}                            & 56.7\%                                       & 31.7\%                                           & 59.7\%                                                    & 53.4\%          &  & 49.0\%         & 32.3\%         & 29.4\%                                                    & 39.6\%          \\
                RT-2-X~\cite{o2024open}                            & 78.7\%                                       & 77.9\%                                           & 25.0\%                                                    & 60.7\%          &  & 82.3\%         & 79.2\%         & 35.3\%                                                    & 64.3\%          \\
                Octo-Base~\cite{team2024octo}                      & 17.0\%                                       & 4.2\%                                            & 22.7\%                                                    & 16.8\%          &  & 0.6\%          & 3.1\%          & 1.1\%                                                     & 1.1\%           \\
                OpenVLA~\cite{kim2024openvla}                      & 16.3\%                                       & 46.2\%                                           & 35.6\%                                                    & 27.7\%          &  & 54.5\%         & 47.7\%         & 17.7\%                                                    & 39.8\%          \\
                RoboVLM (zero-shot)~\cite{li2023generalist}        & 72.7\%                                       & 66.3\%                                           & 26.8\%                                                    & 56.3\%          &  & 68.3\%         & 56.0\%         & 8.5\%                                                     & 46.3\%          \\
                RoboVLM (fine-tuning)~\cite{li2023generalist}      & 77.3\%                                       & 61.7\%                                           & 43.5\%                                                    & 63.4\%          &  & 75.6\%         & 60.0\%         & 10.6\%                                                    & 51.3\%          \\
                $\pi_{0}^\ast$ (BF16 uniform)~\cite{black2024pi_0} & 88.0\%                                       & 80.3\%                                           & 56.0\%                                                    & 70.1\%          &  &                &                &                                                           &                 \\
                \rowcolor[HTML]{EFEFEF} \ours (zero-shot)          & 81.0\%                                       & 69.6\%                                           & 59.3\%                                                    & \textbf{71.9\%} &  & 89.5\%         & 71.7\%         & 36.2\%                                                    & \textbf{68.8\%} \\
                \rowcolor[HTML]{EFEFEF} \ours (fine-tuning)        & 86.0\%                                       & 77.9\%                                           & 57.4\%                                                    & \textbf{75.1\%} &  & 88.0\%         & 72.7\%         & 41.8\%                                                    & \textbf{70.7\%} \\
                \bottomrule
            \end{tabular}
            \begin{tablenotes}
                \footnotesize
                \item $\pi_{0}^\ast$: The results are referred from \href{https://github.com/allenzren/open-pi-zero}{open-pi-zero}.
            \end{tablenotes}
        \end{threeparttable}
    }
    \vspace{-2ex}
\end{table*}

\begin{table*}
  [t]
  \centering
  \caption{\textbf{SimplerEnv evaluation across different policies on WidowX
  Robot tasks}. The zero-shot and fine-tuning results denote the performance of
  OXE dataset~\citep{o2024open} pre-trained models and BridgeData V2~\citep{walke2023bridgedata}
  fine-tuned models, respectively.}
  \resizebox{2\columnwidth}{!}{
  \begin{tabular}{l|cccccccccc}
    \toprule \multirow{2}{*}{Model}               & \multicolumn{2}{c}{\textbf{Put Spoon on Towel}}     & \multicolumn{2}{c}{\textbf{Put Carrot on Plate}} & \multicolumn{2}{c}{\textbf{Stack Green Block on Yellow Block}} & \multicolumn{2}{c}{\textbf{Put Eggplant in Yellow Basket}} & \textbf{\#Overall}                                         \\
                                                  & \begin{tabular}[c]{@{}c@{}}Grasp Spoon\end{tabular} & \begin{tabular}[c]{@{}c@{}}Success\end{tabular}  & \begin{tabular}[c]{@{}c@{}}Grasp Carrot\end{tabular}           & \begin{tabular}[c]{@{}c@{}}Success\end{tabular}            & \begin{tabular}[c]{@{}c@{}}Grasp Green Block\end{tabular} & \begin{tabular}[c]{@{}c@{}}Success\end{tabular} & \begin{tabular}[c]{@{}c@{}}Grasp Eggplant\end{tabular} & \begin{tabular}[c]{@{}c@{}}Success\end{tabular} & Average         \\
    \cmidrule{1-10} RT-1-X~\cite{o2024open}       & 16.7\%                                              & 0\%                                              & 20.8\%                                                         & 4.2\%                                                      & 8.3\%                                                     & 0\%                                             & 0.0\%                                                  & 0\%                                             & 1.1\%           \\
    Octo-Base~\cite{team2024octo}                 & 34.7\%                                              & 12.5\%                                           & 52.8\%                                                         & 8.3\%                                                      & 31.9\%                                                    & 0\%                                             & 66.7\%                                                 & 43.1\%                                          & 16.0\%          \\
    Octo-Small~\cite{team2024octo}                & 77.8\%                                              & 47.2\%                                           & 27.8\%                                                         & 9.7\%                                                      & 40.3\%                                                    & 4.2\%                                           & 87.5\%                                                 & 56.9\%                                          & 30.0\%          \\
    OpenVLA~\cite{kim2024openvla}                 & 4.1\%                                               & 0\%                                              & 33.3\%                                                         & 0\%                                                        & 12.5\%                                                    & 0\%                                             & 8.3\%                                                  & 4.1\%                                           & 1.0\%           \\
    RoboVLM (zero-shot)~\cite{li2023generalist}   & 37.5\%                                              & 20.8\%                                           & 33.3\%                                                         & 25.0\%                                                     & 8.3\%                                                     & 8.3\%                                           & 0.0\%                                                  & 0\%                                             & 13.5\%          \\
    RoboVLM (fine-tuning)~\cite{li2023generalist} & 54.2\%                                              & 29.2\%                                           & 25.0\%                                                         & 25.0\%                                                     & 45.8\%                                                    & 12.5\%                                          & 58.3\%                                                 & 58.3\%                                          & 31.3\%          \\
    \rowcolor[HTML]{EFEFEF} \ours (zero-shot)     & 25.0\%                                              & 20.8\%                                           & 41.7\%                                                         & 20.8\%                                                     & 58.3\%                                                    & 25.0\%                                          & 79.2\%                                                 & 70.8\%                                          & \textbf{34.4\%} \\
    \rowcolor[HTML]{EFEFEF} \ours (fine-tuning)   & 20.8\%                                              & 16.7\%                                           & 29.2\%                                                         & 25.0\%                                                     & 62.5\%                                                    & 29.2\%                                          & 100.0\%                                                & 100.0\%                                         & \textbf{42.7\%} \\
    \bottomrule
  \end{tabular}
  } \label{tab:simplerenv_windowx}
  \vspace{-4ex}
\end{table*}

\begin{figure*}[t]
    \begin{center}
        \includegraphics[width=1\linewidth]{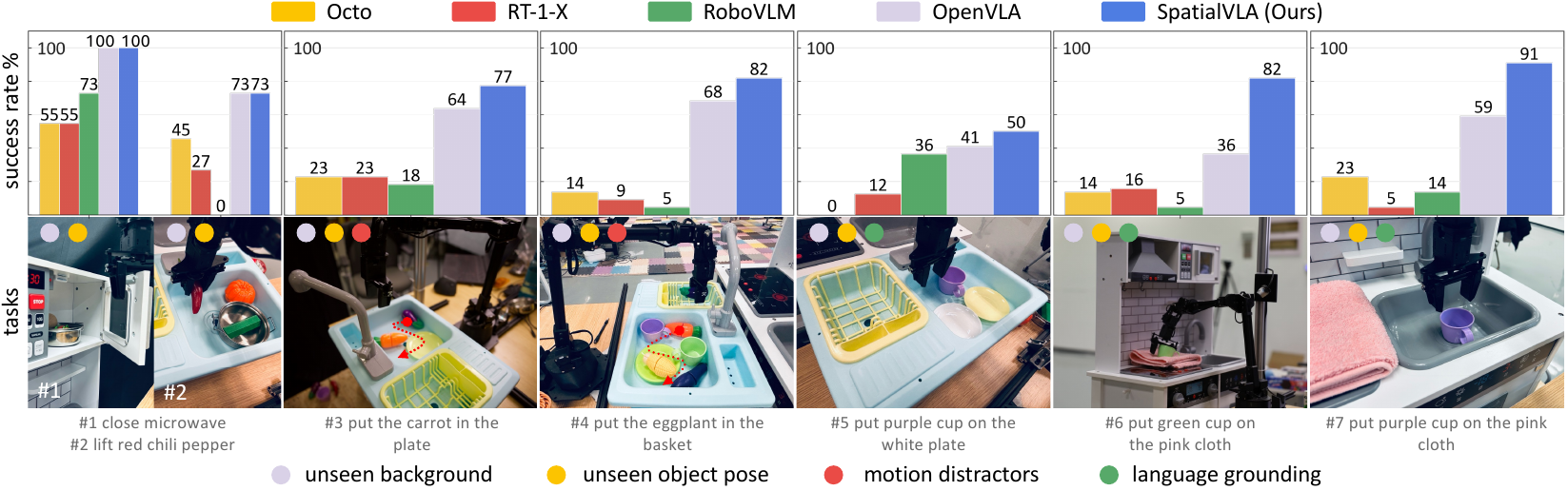}
    \end{center}
    \vspace{-1.5ex}
    \caption{
        \textbf{Zero-shot Robot Control Evaluation on WidowX Robot}. We evaluate \ours across 7 task suites to explore the language grounding, semantic understanding, and motion sensing capabilities, with varying backgrounds, poses, and motion distractors. \ours achieves the highest average success rate, outperforming all generalist manipulation policies.
    }
    \label{fig:widowx_real}
    \vspace{-2.5ex}
\end{figure*}

\subsection{\textbf{Performing Zero-shot Robot Control}} 
\label{sec:zero_shot_robot_control}
\label{sec:zero_shot_robot_control}
\textbf{Evaluation Setups and Baselines.} To assess the robustness of \ours in diverse environmental variations, we employ the SimplerEnv simulation benchmark~\citep{li24simpler} to evaluate visual matching and variant aggregation metrics. SimplerEnv features WidowX and Google Robot setups, providing diverse manipulation scenarios with varied lighting, color, textures, and robot camera pose conditions, bridging the visual appearance gap between real and simulated environments. We compare our model with the latest state-of-the-art generalist manipulation policies, including RT-1~\citep{brohan2022rt}, RT-1-X~\cite{o2024open}, RT-2-X~\cite{o2024open}, Octo~\cite{team2024octo}, OpenVLA~\cite{kim2024openvla}, HPT~\citep{wang2024scaling}, TraceVLA~\cite{zheng2024tracevla}, and RoboVLM~\cite{li2023generalist}. Where RT-1-X, RT-2-X, Octo, OpenVLA, HPT, TraceVLA, and RoboVLM are trained with mixtures of OXE dataset~\citep{o2024open}. Since RT-1 is trained with the Google Fractal Dataset~\citep{brohan2022rt}, we also compare RT-1 with our method fine-tuned on the Google Fractal and BridgeData V2~\citep{walke2023bridgedata}.

For a more comprehensive evaluation, we conduct experiments on a real-world WidowX robot platform from the BridgeData V2 evaluation~\citep{walke2023bridgedata}. As shown in~\cref{fig:widowx_real}, we design seven task suites for the WidowX robot, encompassing \textbf{language grounding}, \textbf{semantic understanding} (unseen background and poses), and \textbf{motion distractors} (manually move the object). All generalist manipulation policies, including Octo, RT-1-X, OpenVLA, and RoboVLM, are evaluated across 7 task suites with 11 trials each, resulting in a total of 77 rollouts. A more detailed breakdown of all tasks and policy settings can be found in the~\cref{supp:eval_setup}. 

\textbf{SimplerEnv Evaluation of Google Robot and WidowX.} \cref{tab:simplerenv_google_robot} summarizes the zero-shot and fine-tuning results across different manipulation policies on the Google Robot setup. On average, \ours achieves the highest overall visual matching and variant aggregation performance with a significant margin. Our \ours model yields 71.9\% and 75.1\% Visual Matching scores in zero-shot and fine-tuning settings, surpassing the second-best policy, RoboVLM, by +15.6\% and +11.7\% margins. Notably, our model trained from scratch on OXE mixture with RH20T surpasses the state-of-the-art closed-source model RT-2-X~\cite{o2024open}, achieving superior performance in Visual Matching (71.9\% vs 60.7\%) and Variant Aggregation (68.8\% vs 64.3\%), while using significantly fewer model parameters (3.5B vs 55B). Qualitatively, we find that \ours exhibits greater generalizability and robustness across diverse robotic manipulation tasks and environmental conditions, characterized by varying visual appearances, which is further supported by its superior performance in variant aggregation. In particular, \ours also matches or outperforms the latest SOTA model $\pi_0$. \cref{tab:simplerenv_windowx} summarizes the results across different manipulation policies on the WidowX setup. Our model surpasses the state-of-the-art RoboVLM policy, achieving overall success rates of 34.4\% and 42.7\%. Fine-tuning on the BridgeV2 yields a remarkable 100\% success rate in the ``Put Eggplant in Yellow Basket" task, demonstrating the model's exceptional zero-shot manipulation capability.

\begin{figure*}[t]
    \vspace{-2ex}
    \begin{center}
        \includegraphics[width=1\linewidth]{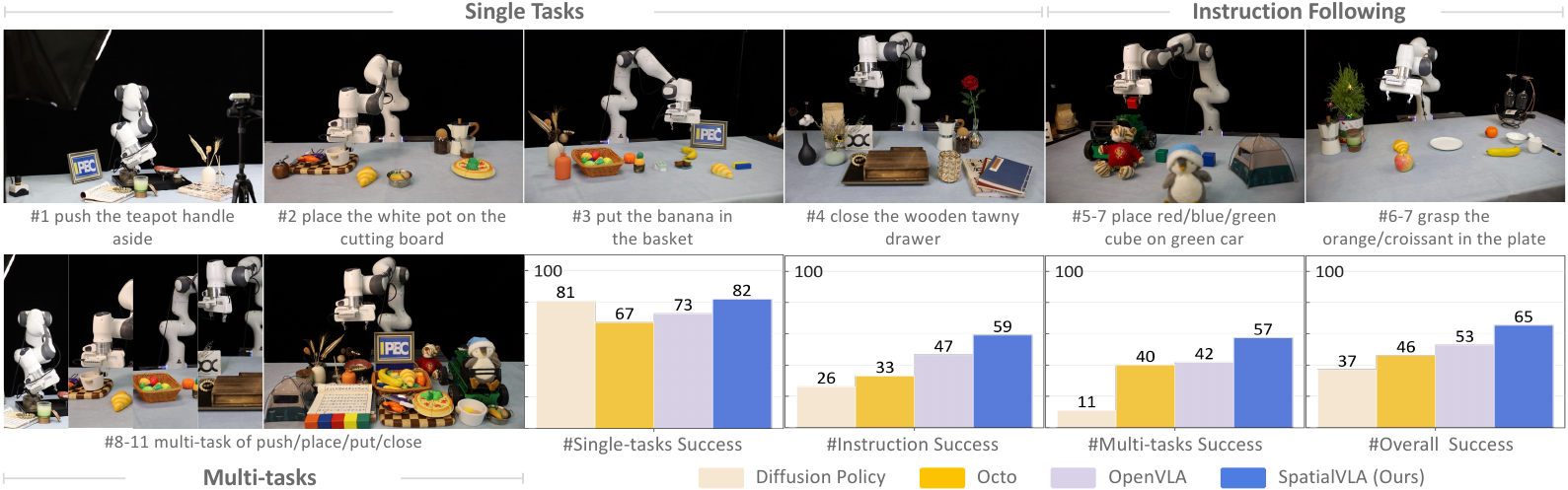}
    \end{center}
    \vspace{-1.5ex}
    \caption{
        \textbf{Adapting to New Robot Setups on Franka Robot}. \ours serves as a generalist robot control policy, achieving better performance across multiple setups, and can be effectively used as an initialization for fine-tuning to new robot tasks.
    }
    \label{fig:franka_robo_sft}
\end{figure*}
\begin{table*}[h]
    \centering
    \caption{\textbf{LIBERO Simulation Benchmark Results.} We present the success rate (SR) and standard error for each method across four task suites, which are averaged over three random seeds with 500 trials. Fine-tuned \ours models achieve the highest average success rate and ranking, followed by fine-tuned OpenVLA~\citep{kim2024openvla} and Octo~\citep{team2024octo}.}
    \label{tab:libero} 
    \resizebox{\textwidth}{!}{\begin{tabular}{l|cc|cc|cc|cc|cc}
        \toprule                                                      & \multicolumn{2}{c|}{LIBERO-Spatial} & \multicolumn{2}{c|}{LIBERO-Object} & \multicolumn{2}{c|}{LIBERO-Goal} & \multicolumn{2}{c|}{LIBERO-Long} & \multicolumn{2}{c}{Average} \\
        \cline{2-3} \cline{4-5} \cline{6-7} \cline{8-9} \cline{10-11} & SR ($\uparrow$)                     & Rank ($\downarrow$)                & SR ($\uparrow$)                  & Rank ($\downarrow$)              & SR ($\uparrow$)            & Rank ($\downarrow$) & SR ($\uparrow$)           & Rank ($\downarrow$) & SR ($\uparrow$)           & Rank ($\downarrow$) \\
        \midrule Diffusion Policy from scratch                        & 78.3 $\pm$ 1.1\%                    & 5                                  & \textbf{92.5 $\pm$ 0.7\%}        & 1                                & 68.3 $\pm$ 1.2\%           & 5                   & 50.5 $\pm$ 1.3\%          & 5                   & 72.4 $\pm$ 0.7\%          & 5                   \\
        \midrule Octo fine-tuned                                      & 78.9 $\pm$ 1.0\%                    & 4                                  & 85.7 $\pm$ 0.9\%                 & 4                                & \textbf{84.6 $\pm$ 0.9\%}  & 1                   & 51.1 $\pm$ 1.3\%          & 4                   & 75.1 $\pm$ 0.6\%          & 3                   \\
        OpenVLA fine-tuned                                            & 84.7 $\pm$ 0.9\%                    & 2                                  & 88.4 $\pm$ 0.8\%                 & 3                                & 79.2 $\pm$ 1.0\%           & 2                   & 53.7 $\pm$ 1.3\%          & 3                   & 76.5 $\pm$ 0.6\%          & 2                   \\
        TraceVLA fine-tuned                                           & 84.6 $\pm$ 0.2\%                    & 3                                  & 85.2 $\pm$ 0.4\%                 & 5                                & 75.1 $\pm$ 0.3\%           & 4                   & 54.1 $\pm$ 1.0\%          & 2                   & 74.8 $\pm$ 0.5\%          & 4                   \\
        \ours fine-tuned                                              & \textbf{88.2 $\pm$ 0.5\%}           & 1                                  & \textbf{89.9 $\pm$ 0.7\%}        & 2                                & 78.6 $\pm$ 0.6\%           & 3                   & \textbf{55.5 $\pm$ 1.0}\% & 1                   & \textbf{78.1 $\pm$ 0.7\%} & 1                   \\
        \bottomrule
    \end{tabular}}
\end{table*}

\textbf{Real-world WidowX Evaluation.} \cref{fig:widowx_real} presents the results of the real-world ``out-of-the-box" evaluation in WidowX robot platform. We observe that, in simple single-task scenarios (\#1 close microwave), all the policies exhibit some generalizability, successfully completing tasks in unseen environments. However, in moderately complex tasks (\#3-7), most policies, such as RT-1-X, Octo, and RoboVLM struggle with manipulation, frequently encountering issues like object misidentification and grasp failures. Compared to OpenVLA, our method demonstrates superior robustness in handling motion disturbances (human-induced dynamic object movement in tasks \#3 and \#4), successfully tracking and grasping carrot and eggplant. Furthermore, in the instruction-following tasks (\#5-7), our method demonstrates strong instruction-following ability, accurately executing tasks like picking up a green cup and placing it on a white plate, not a pink one, based on color descriptions in the prompts, outperforming OpenVLA and other generalist policies. Overall, \ours achieves a higher average success rate, showcasing robust and generalizable operation capabilities in unseen scenarios, objects, language grounding, and dynamic motions.

\subsection{\textbf{Adapting to New Robot Setups}}
\label{sec:new_robot_setup}
\noindent \textbf{Evaluation Setup and Comparisons.} We present the evaluation of \ours on the LIBERO simulation benchmark~\citep{liu2023libero}, which consists of a set of diverse robotic manipulation tasks in simulated environments.  Following OpenVLA~\citep{kim2024openvla}, we conduct experiments on four task suites, each comprising 10 tasks with 50 human-teleoperated demonstrations. These suites evaluate the model's understanding of spatial relationships (\textbf{LIBERO-Spatial}), object types (\textbf{LIBERO-Object}), task-oriented behaviors (\textbf{LIBERO-Goal}), and its ability to generalize to long-horizon tasks with diverse objects, layouts, and goals (\textbf{LIBERO-Long}). We compare our approach to several generalist manipulation policy methods, including Diffusion Policy~\citep{chi2024diffusionpolicy}, Octo~\citep{team2024octo}, OpenVLA~\citep{kim2024openvla}, and TraceVLA~\citep{zheng2024tracevla}. \ours is fine-tuned on the corresponding dataset for 200 epochs using LoRA ($r=32$, $\alpha=32$), which incorporates spatial embedding adaption in~\cref{sec:training} from new Gaussian distribution. 

To facilitate a more comprehensive evaluation, 13 Franka tasks are established to validate the model's manipulation performance, as shown in~\cref{fig:franka_robo_sft}. The evaluation consists of three setups: \textbf{Single Task}, which includes four basic tasks: pick, place, push, and close; \textbf{Instruction Following}, which involves manipulating different objects in the same scene based on language instructions; and \textbf{Multi-tasks}, which involves training on a mixture of all four single-task data and tested on these tasks. We compare \ours with mainstream policies, including Diffusion Policy, Octo, and OpenVLA. More details can be found in the~\cref{supp:eval_setup}.

\noindent \textbf{Evaluation Results.} \cref{tab:libero} present the LIBERO~\citep{liu2023libero} experimental results. Notably, we observe that \ours can be effectively adapted to tasks in the LIBERO environments, as it obtains the highest average success rate of 78.1\% and the first rank across all the policies. In particular, \ours achieves a remarkable 88.2\% success rate on the LIBERO-Spatial task, which consists of different object layouts, demonstrating the model's strong understanding of spatial relationships. In most tasks, \ours outperforms the state-of-the-art generalist manipulation policies but struggles with long-horizon tasks in \textbf{LIBERO-Long}, due to the lack of architecture design for long-horizon observation.

\cref{fig:franka_robo_sft} summarizes the results of the Franka robot fine-tuning evaluation. In single-task tests, \ours and Diffusion Policy show similar accuracy (82\% vs 81\%), outperforming OpenVLA and Octo. However, in the instruction following tasks, \ours improves by +12\% over OpenVLA, while Diffusion Policy struggles with a 26\% success rate. In multi-tasks, \ours leverages its pre-training on OXE and 3D perception capabilities to achieve a 57\% accuracy rate, surpassing other generalist policies. In summary, \ours demonstrates its versatility as a generalist robot control policy, achieving better performance across various tasks, and can be effectively used as an initialization for new robot tuning.

\begin{figure}[t]
    \begin{center}
        \includegraphics[width=1\linewidth]{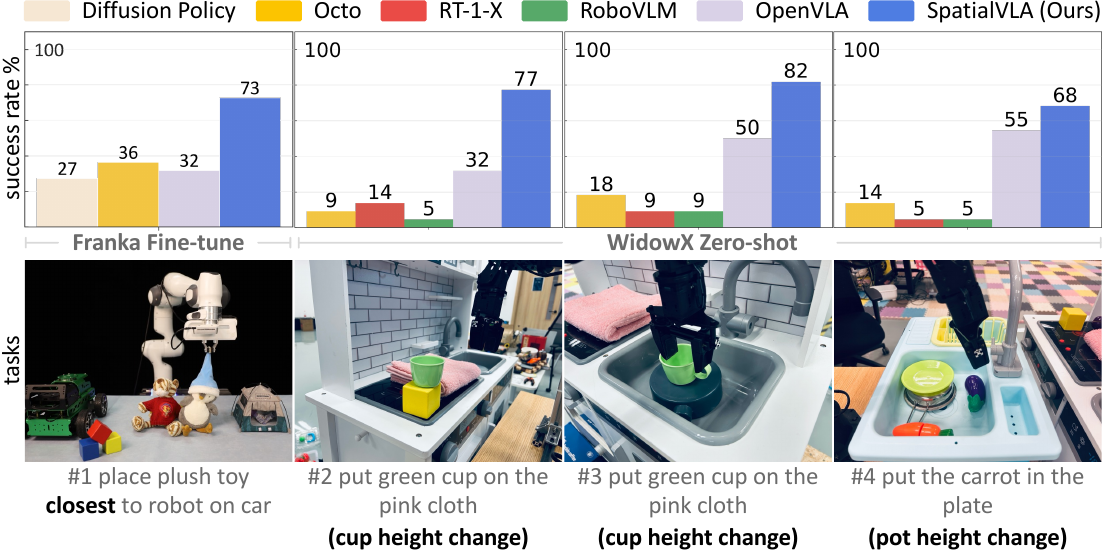}
    \end{center}
    \vspace{-1.5ex}
    \caption{\textbf{Spatial Understanding Capability Evaluation}. Benefiting from the proposed Ego3D Position Encoding, \ours exhibits superior performance in understanding spatial prompts and complex spatial layout tasks.}
    \label{fig:spatial_setup}
    \vspace{-2.5ex}
\end{figure}

\subsection{\textbf{Evaluating Spatial Understanding Capability}} 
\label{sec:evaluating_spatial_understanding_capability}
\noindent \textbf{Evaluation Setup and Comparisons.} As shown in~\cref{fig:spatial_setup,tab:libero}, we evaluate the spatial understanding capabilities of \ours through on three robot setups: \textbf{Franka Robot} fine-tuning, \textbf{WidowX Robot} zero-shot, and \textbf{Libero-Spatial} fine-tuning. The tasks exhibit varying spatial complexities, with the Franka task involving prompt understanding (e.g., \#1 place plush toy \textbf{closest to} robot on car), the WidowX task featuring explicit \textbf{height changes} (e.g., \#2 put green cup on the pink cloth), and the LIBERO-Spatial task involving object layout variations. Seven mainstream policies, namely Diffusion Policy, Octo, RT-1-X, OpenVLA, TraceVLA, and RoboVLM, are employed for comparison.

\noindent \textbf{Evaluation Results.} Compared to existing policies, \ours shows superior spatial understanding, achieving 73\% accuracy in Franka task \#1, which involves spatial prompts, and significantly improving manipulation capabilities for complex positional changes in the out-of-distribution WidowX Zero-shot tasks \#2-4. Similar results are observed in the LIBERO-Spatial task suite (88.2\% success rate). Policies like Octo, Diffusion Policy, and OpenVLA, which lack integrated depth information, face significant challenges in adapting to spatial layout changes, yielding a success rate consistently lower than 50\%. Consequently, we suggest integrating 3D information (\cref{sec:method}), including depth or point cloud, into the VLA framework to improve the model's adaptability and robustness in spatial layout variations.

\begin{table*}
    [t]
    \centering
    \caption{\textbf{Pre-training Ablations on the Mixture Dataset of \textcolor{cvprblue}{Google Fractal} and \textcolor{orange}{BridgeData V2}}. Initializing a high-resolution action grid from
        the data distribution and 3D position encoding enhances the model’s
        generalization capability.}
    \resizebox{2\columnwidth}{!}{
        \begin{tabular}{ll|ccccccccccc}
            \toprule 
            & \multirow{3}{*}{\#setting} & \multicolumn{2}{c}{\textcolor{cvprblue}{Pick Coke Can}} &                     & \multicolumn{2}{c}{\textcolor{cvprblue}{Move Near}} &  & \multicolumn{2}{c}{\textcolor{orange}{Put Carrot on Plate}} &                 & \multicolumn{2}{c}{\textcolor{orange}{Put Eggplant in Yellow Basket}}                                                        \\
            \cmidrule{3-4} \cmidrule{6-7} \cmidrule{9-10} \cmidrule{12-13}
                                                &                                                         & variant aggregation & visual matching                                     &  & variant aggregation                                         & visual matching &                                                                       & grasp carrot & success &  & grasp eggplant & success \\
            \cmidrule{1-4} \cmidrule{6-7} \cmidrule{9-10} \cmidrule{12-13}
            \#All                               & $[1]$. \textcolor{cvprblue}{\textbf{\ours}}             & 81.6\%              & 70.7\%                                              &  & 79.2\%                                                      & 85.4\%          &                                                                       & 41.7\%       & 33.3\%  &  & 91.7\%         & 87.5\%  \\
            \hline
            Linear Discretization               & $[2]$. $\sim$ linear 256 bins                           & 40.7\%              & 19.0\%                                              &  & 47.1\%                                                      & 52.9\%          &                                                                       & 41.7\%       & 33.3\%  &  & 87.5\%         & 70.8\%  \\
            Distribution                        & $[3]$. $\sim$ uniform distribution                      & 77.9\%              & 28.0\%                                              &  & 64.2\%                                                      & 55.0\%          &                                                                       & 45.8\%       & 12.5\%  &  & 79.2\%         & 54.2\%  \\
            \hline
                                                & $[4]$. resolution 1026                                  & 74.4\%              & 67.3\%                                              &  & 59.1\%                                                      & 54.2\%          &                                                                       & 45.8\%       & 25.0\%  &  & 66.7\%         & 54.2\%  \\
            Action Grids                        & $[5]$. resolution 4610                                  & 76.7\%              & 68.0\%                                              &  & 69.8\%                                                      & 79.2\%          &                                                                       & 41.7\%       & 33.3\%  &  & 83.3\%         & 75.0\%  \\
            Resolution                          & $[6]$. resolution 6166                                  & 80.9\%              & 74.0\%                                              &  & 74.0\%                                                      & 79.2\%          &                                                                       & 41.7\%       & 33.3\%  &  & 95.8\%         & 87.5\%  \\
                                                & $[7]$. resolution 8194                                  & 81.6\%              & 70.7\%                                              &  & 79.2\%                                                      & 85.4\%          &                                                                       & 41.7\%       & 33.3\%  &  & 91.7\%         & 87.5\%  \\
            \hline
            \multirow{2}{*}{Encoding}           & $[8]$. $-$ ego3d encoding                               & 68.9\%              & 70.3\%                                              &  & 66.7\%                                                      & 62.0\%          &                                                                       & 54.2\%       & 12.5\%  &  & 75.0\%         & 37.5\%  \\
                                                & $[9]$. $-$ freeze llm embedding                         & 70.2\%              & 50.7\%                                              &  & 63.1\%                                                      & 62.5\%          &                                                                       & 33.3\%       & 20.8\%  &  & 95.8\%         & 79.2\%  \\
            \bottomrule
        \end{tabular}
    } \label{tab:ablation_pretrain}
\end{table*}
\begin{table*}
    [t]
    \centering
    \caption{\textbf{Fine-tuning Ablations in Domain Datasets}. Pretrained
    models are full parameter fine-tuned in individual \textcolor{cvprblue}{Google Fractual} and \textcolor{orange}{Bridge
    V2} Dataset. In LIBERO tasks, both full-tuning and LoRA-tuning are applied. fine-tuned
    with Gaussian adaptation from new dataset distribution helps align spatial grid
    features and improve initialization and accelerating convergence.}
    \resizebox{2\columnwidth}{!}{
    \begin{tabular}{l|ccccccccccc}
        \toprule \multirow{2}{*}{\#setting}                                                     & \multicolumn{2}{c}{\textcolor{cvprblue}{Pick Coke Can}} &                 & \multicolumn{2}{c}{\textcolor{cvprblue}{Move Near}} &                     & \multicolumn{2}{c}{\textcolor{orange}{Put Carrot on Plate}} &  & \multicolumn{2}{c}{\textcolor{orange}{Put Eggplant in Yellow Basket}} \\
        \cmidrule{2-3} \cmidrule{5-6} \cmidrule{8-9} \cmidrule{11-12}                           & variant aggregation                                     & visual matching &                                                     & variant aggregation & visual matching                                             &  & grasp carrot                                                         & success &  & grasp eggplant & success \\
        \cmidrule{1-3} \cmidrule{5-6} \cmidrule{8-9} \cmidrule{11-12} $[1]$. full params tuning & 88.0\%                                                  & 77.0\%          &                                                     & 72.7\%              & 75.0\%                                                      &  & 29.2\%                                                               & 20.8\%  &  & 100\%          & 91.7\%  \\
        $[2]$. $+$ Gaussian adaption                                                            & 90.1\%                                                  & 86.0\%          &                                                     & 74.6\%              & 77.9\%                                                      &  & 29.2\%                                                               & 25.0\%  &  & 100\%          & 100\%   \\
        \midrule \midrule \#setting                                                             & \multicolumn{2}{c}{LIBERO-Spatial}                      &                 & \multicolumn{2}{c}{LIBERO-Object}                   &                     & \multicolumn{2}{c}{LIBERO-Goal}                             &  & \multicolumn{2}{c}{LIBERO-Long}                                       \\
        \cmidrule{1-3} \cmidrule{5-6} \cmidrule{8-9} \cmidrule{11-12} $[3]$. Full params tuning & \multicolumn{2}{c}{77.7 $\pm$ 0.4\%}                    &                 & \multicolumn{2}{c}{73.3 $\pm$ 0.4\%}                &                     & \multicolumn{2}{c}{78.5 $\pm$ 0.5\%}                        &  & \multicolumn{2}{c}{43.9 $\pm$ 0.8\%}                                  \\
        $[4]$. LoRA tuning                                                                      & \multicolumn{2}{c}{83.6 $\pm$ 0.7\%}                    &                 & \multicolumn{2}{c}{84.8 $\pm$ 0.9\%}                &                     & \multicolumn{2}{c}{76.4 $\pm$ 0.2\%}                        &  & \multicolumn{2}{c}{50.1 $\pm$ 0.3\%}                                  \\
        $[5]$. $+$ Spatial embedding adaption                                                   & \multicolumn{2}{c}{\textbf{88.2 $\pm$ 0.5\%}}           &                 & \multicolumn{2}{c}{\textbf{89.9 $\pm$ 0.7\%}}       &                     & \multicolumn{2}{c}{\textbf{78.6 $\pm$ 0.6\%}}               &  & \multicolumn{2}{c}{\textbf{55.5 $\pm$ 1.0\%}}                         \\
        \bottomrule
    \end{tabular}
    } \label{tab:ablation_fintuning}
\end{table*}

\subsection{\textbf{Ablations on Design  Decisions}}
\label{sec:ablation}
In this section, we conduct ablation studies to investigate the effectiveness of the proposed 3D Spatial Presentation in both \textit{pre-training} and \textit{post-training} stages. 
\begin{figure}[t]
    \begin{center}
        \includegraphics[width=0.98\linewidth]{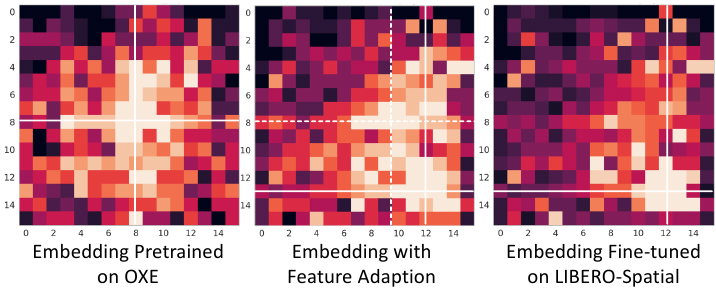}
    \end{center}
    \vspace{-1.5ex}
    \caption{Cross-sectional features visualization in spatial grids. The proposed spatial embedding adaptation aligns the pre-trained spatial grid features with those of the target fine-tuned model, improving initialization and accelerating convergence.}
    \label{fig:emb_adpt}
    \vspace{-2.5ex}
\end{figure}

\noindent \textbf{Pre-training in Mixture Dataset.} The pre-training ablations in~\cref{tab:ablation_pretrain} are conducted on a mixture dataset that combines Google Fractal~\citep{brohan2022rt} and BridgeData V2~\citep{walke2023bridgedata}. All the models are trained from scratch on 8 A100 GPUs with 128 batch size for 120k steps. We select four tasks from the SimplerEnv benchmark~\citep{li24simpler}, namely ``Pick Coke Can" and ``Move Near" on the Google Robot, as well as ``Put Carrot on Plate" and ``Put Eggplant in Yellow Basket" on the WidowX Robot, to dissect the model's component-wise performance. 

In contrast to the conventional linear 256-bin action space discretization~\citep{brohan2022rt,o2024open,kim2024openvla} (\#1\textit{v.s.}\#2), the proposed \textit{adaptive spatial action grids exhibits significant advantages}, particularly in the Google Robot task, with the promotion of +36.5\% and +42.1\% in variant aggregation and visual matching success rates, respectively. During model training, we also observe that models using linear 256-bin discretization converge slower, despite achieving lower L1 Loss. Another suggestion is to \textit{initialize the grid partitioning based on the dataset distribution}, rather than using a uniform grid (\#1\textit{v.s.}\#3), which enables the model to focus on high-frequency action spaces adaptively and further improves its generalization capabilities.

Compared to 1026-resolution action grids (\#1\textit{v.s.}\#4), where $\text{M}_\text{trans}=\text{M}_\text{trans}=512, \text{M}_\text{grip}=2$, \ours with 8194-resolution action grids ($\text{M}_\text{trans}=\text{M}_\text{trans}=4096, \text{M}_\text{grip}=2$) achieves significant performance boosts, particularly in ``move near" and ``put eggplant in yellow basket" tasks, with success rate increments of +31.2\% and +33.3\%. Additionally, we find that \textit{lower-resolution models tend to learn smaller actions, causing slow motion issues, and high-resolution models exhibit improved transfer performance in the fine-tuning stage}. 

According to the ablation results (\#1\textit{v.s.}\#8), the proposed egocentric 3D position encoding (ego3d), \textit{incorporating 3D point cloud features, helps the model overcome varied lighting, color, textures, and camera poses}, yielding stronger generalizability in diverse manipulation scenarios. Models w/o ego3d suffer a significant performance drop in variant aggregation, from 81.6\% and 79.2\% to 68.9\% and 66.7\%, due to their inability to adapt to scene changes. During pre-training, we also observe from (\#1\textit{v.s.}\#9) that \textit{freezing the language embedding and sharing a trainable spatial embedding helps to improve the model's manipulation capabilities}, which is also beneficial for faster training and instruction following. 

\noindent \textbf{Post-training in Domain Dataset.} We conduct post-training ablations in~\cref{tab:ablation_fintuning}, separately fine-tuning on large-scale datasets Google Fractal and BridgeData V2 and BridgeData V2, and comparing full fine-tuning and LoRA-tuning on the small-scale LIBERO datasets~\citep{liu2023libero}. The spatial embedding adaption denotes partitioning spatial grids from the new dataset Gaussian distribution and updating the spatial feature embedding with the grids. 

On large-scale datasets (\#1\textit{v.s.}\#2), models fine-tuned with spatial embedding adaptation yield marginal gains of +2.9\% in visual matching on Move Near), as the large-scale dataset distribution closely matches the pre-training distribution, allowing the model to learn fine-grained features thereby limiting the benefits of the adaption. While, on the LIBERO small dataset tasks (\#4\textit{v.s.}\#5), initializing the feature grid with the new distribution boosts model performance by +4.6\%, +5.1\%, +2.2\%, and +5.4\% on LIBERO-Spatial, LIBERO-Object, LIBERO-Goal, and LIBERO-Long, respectively. As shown in~\cref{fig:emb_adpt}, \textit{\textit{feature adaptation from the new distribution aligns pre-trained spatial features with the target fine-tuned model}}, improving initialization and accelerating convergence. Moreover, LoRA fine-tuning outperforms full-parameter fine-tuning on small dataset tasks (\#3\textit{v.s.}\#4), making LoRA the preferred method for small datasets.
\section{Discussion, Limitations, and Future Work}
\label{sec:conclusion}
In this paper, we present \ours, an innovative vision-language-action model to explore efficient spatial representations for generalist robot policy. \ours introduces Ego3D position encoding and adaptive action grids to inject 3D awareness into robot observation representation and spatial action tokenization through robot-agnostic designs, equipping the VLA models with the spatial understanding ability of the 3D physical world. After pre-training on large-scale heterogeneous robot datasets, we find that \ours is a more generalizable and transferrable generalist policy for zero-shot robot control. Our extensive real-world and simulated robot experiments show that \ours leads to dramatically improved performance over the previous VLA models, especially on tasks that require precise spatial understanding. We also show that the pre-trained \ours model can effectively adapt to new robot setups and tasks via action grids re-discretization, which offers a new way for robot-specific post-training. In the following, we discuss our limitations of \ours and potential solutions, hoping to inspire further innovative works.

\boldparagraph{More Generalizable Distribution Fitting.} In this paper, \ours fits action signals with Gaussian distributions to encode actions as spatial grids, demonstrating remarkable generalizability and flexible adaptation to new robot setups through re-initialized grids and token embeddings. However, this raises a crucial question: Is modeling data distributions as Gaussian optimal? We argue that Gaussian modeling is suboptimal, as it can lead to grid clustering on specific coordinate axes in extreme robot operation scenarios, such as single-axis motion, resulting in lost motion capabilities on other axes. Moreover, dataset noises can further distort the spatial grid distribution. One future solution is to combine implicit data distribution modeling techniques, such as Variational Auto-Encoder-based high-dimensional feature space mapping, with explicit grid partitioning, enhancing action presentation efficiency and noise robustness.

\boldparagraph{More Flexible VLA architectures.} In our implementation, we predict spatial action tokens through the autoregressive paradigm and further decode them into actions, resulting in each action being represented by 3 tokens. Although \ours achieves 21Hz inference speed, it is slower than diffusion decoding~\citep{black2024pi_0,chi2024diffusionpolicy,li2023generalist}, which decodes tokens into multiple consecutive actions. In the future, integrating diffusion decoding with spatial grid action presentation and exploring dynamic token numbers for action mapping will be valuable. Furthermore, as the model relies solely on current frame observations and history tokens for action prediction, it faces challenges in long-horizon tasks, similar to other generalizable policies~\citep{chi2024diffusionpolicy,team2024octo}. Future work should focus on designing efficient historical information perception mechanisms to enhance the model's long-sequence modeling capabilities, enabling seamless task switching in real-time manipulation scenarios.

\boldparagraph{Higher-Quality Diverse Data.} \ours is pre-trained on OXE and RH20T, but the variable quality of OXE data can hinder training. Therefore, future work exploring optimal data composition and distilling high-quality subsets from the heterogeneous robot data collections is vital for boosting model efficiency and generalizability.

{\small

\boldparagraph{Acknowledgements.}
This work is supported by the Shanghai AI Laboratory and the National Natural Science Foundation of China (624B2044).
}

\bibliographystyle{plainnat}
\bibliography{main}

\begin{thebibliography}{77}
\providecommand{\natexlab}[1]{#1}
\providecommand{\url}[1]{\texttt{#1}}
\expandafter\ifx\csname urlstyle\endcsname\relax
  \providecommand{\doi}[1]{doi: #1}\else
  \providecommand{\doi}{doi: \begingroup \urlstyle{rm}\Url}\fi

\bibitem[Alayrac et~al.(2022)Alayrac, Donahue, Luc, Miech, Barr, Hasson, Lenc, Mensch, Millican, Reynolds, et~al.]{alayrac2022flamingo}
Jean-Baptiste Alayrac, Jeff Donahue, Pauline Luc, Antoine Miech, Iain Barr, Yana Hasson, Karel Lenc, Arthur Mensch, Katherine Millican, Malcolm Reynolds, et~al.
\newblock Flamingo: a visual language model for few-shot learning.
\newblock In \emph{Proceedings of the Conference on Neural Information Processing System (NeurIPS)}, 2022.

\bibitem[Belkhale et~al.(2023)Belkhale, Cui, and Sadigh]{belkhale2023hydra}
Suneel Belkhale, Yuchen Cui, and Dorsa Sadigh.
\newblock Hydra: Hybrid robot actions for imitation learning.
\newblock In \emph{Proceedings of the Conference on Robot Learning (CoRL)}, 2023.

\bibitem[Bharadhwaj et~al.(2024)Bharadhwaj, Vakil, Sharma, Gupta, Tulsiani, and Kumar]{bharadhwaj2024roboagent}
Homanga Bharadhwaj, Jay Vakil, Mohit Sharma, Abhinav Gupta, Shubham Tulsiani, and Vikash Kumar.
\newblock Roboagent: Generalization and efficiency in robot manipulation via semantic augmentations and action chunking.
\newblock In \emph{Proceedings of the IEEE International Conference on Robotics and Automation (ICRA)}, 2024.

\bibitem[Bhat et~al.(2023)Bhat, Birkl, Wofk, Wonka, and M{\"u}ller]{bhat2023zoedepth}
Shariq~Farooq Bhat, Reiner Birkl, Diana Wofk, Peter Wonka, and Matthias M{\"u}ller.
\newblock Zoedepth: Zero-shot transfer by combining relative and metric depth.
\newblock \emph{arXiv preprint arXiv:2302.12288}, 2023.

\bibitem[Black et~al.(2024)Black, Brown, Driess, Esmail, Equi, Finn, Fusai, Groom, Hausman, Ichter, et~al.]{black2024pi_0}
Kevin Black, Noah Brown, Danny Driess, Adnan Esmail, Michael Equi, Chelsea Finn, Niccolo Fusai, Lachy Groom, Karol Hausman, Brian Ichter, et~al.
\newblock $\pi_0$: A vision-language-action flow model for general robot control.
\newblock \emph{arXiv preprint arXiv:2410.24164}, 2024.

\bibitem[Brohan et~al.(2022)Brohan, Brown, Carbajal, Chebotar, Dabis, Finn, Gopalakrishnan, Hausman, Herzog, Hsu, et~al.]{brohan2022rt}
Anthony Brohan, Noah Brown, Justice Carbajal, Yevgen Chebotar, Joseph Dabis, Chelsea Finn, Keerthana Gopalakrishnan, Karol Hausman, Alex Herzog, Jasmine Hsu, et~al.
\newblock Rt-1: Robotics transformer for real-world control at scale.
\newblock \emph{arXiv preprint arXiv:2212.06817}, 2022.

\bibitem[Brohan et~al.(2023)Brohan, Brown, Carbajal, Chebotar, Chen, Choromanski, Ding, Driess, Dubey, Finn, et~al.]{brohan2023rt}
Anthony Brohan, Noah Brown, Justice Carbajal, Yevgen Chebotar, Xi~Chen, Krzysztof Choromanski, Tianli Ding, Danny Driess, Avinava Dubey, Chelsea Finn, et~al.
\newblock Rt-2: Vision-language-action models transfer web knowledge to robotic control.
\newblock \emph{arXiv preprint arXiv:2307.15818}, 2023.

\bibitem[Cheang et~al.(2024)Cheang, Chen, Jing, Kong, Li, Li, Liu, Wu, Xu, Yang, et~al.]{cheang2024gr}
Chi-Lam Cheang, Guangzeng Chen, Ya~Jing, Tao Kong, Hang Li, Yifeng Li, Yuxiao Liu, Hongtao Wu, Jiafeng Xu, Yichu Yang, et~al.
\newblock Gr-2: A generative video-language-action model with web-scale knowledge for robot manipulation.
\newblock \emph{arXiv preprint arXiv:2410.06158}, 2024.

\bibitem[Chen et~al.()Chen, Adebola, and Goldberg]{BerkeleyUR5Website}
Lawrence~Yunliang Chen, Simeon Adebola, and Ken Goldberg.
\newblock Berkeley {UR5} demonstration dataset.
\newblock \url{https://sites.google.com/view/berkeley-ur5/home}.

\bibitem[Chen et~al.(2024{\natexlab{a}})Chen, Chen, Zhang, Li, Yu, Fei, Zhu, Fan, and Chen]{chen2024ll3da}
Sijin Chen, Xin Chen, Chi Zhang, Mingsheng Li, Gang Yu, Hao Fei, Hongyuan Zhu, Jiayuan Fan, and Tao Chen.
\newblock Ll3da: Visual interactive instruction tuning for omni-3d understanding reasoning and planning.
\newblock In \emph{Proceedings of the IEEE/CVF Conference on Computer Vision and Pattern Recognition (CVPR)}, 2024{\natexlab{a}}.

\bibitem[Chen et~al.(2024{\natexlab{b}})Chen, Djolonga, Padlewski, Mustafa, Changpinyo, Wu, Ruiz, Goodman, Wang, Tay, et~al.]{chen2023pali}
Xi~Chen, Josip Djolonga, Piotr Padlewski, Basil Mustafa, Soravit Changpinyo, Jialin Wu, Carlos~Riquelme Ruiz, Sebastian Goodman, Xiao Wang, Yi~Tay, et~al.
\newblock Pali-x: On scaling up a multilingual vision and language model.
\newblock In \emph{Proceedings of the IEEE/CVF Conference on Computer Vision and Pattern Recognition (CVPR)}, 2024{\natexlab{b}}.

\bibitem[Chi et~al.(2023)Chi, Xu, Feng, Cousineau, Du, Burchfiel, Tedrake, and Song]{chi2024diffusionpolicy}
Cheng Chi, Zhenjia Xu, Siyuan Feng, Eric Cousineau, Yilun Du, Benjamin Burchfiel, Russ Tedrake, and Shuran Song.
\newblock Diffusion policy: Visuomotor policy learning via action diffusion.
\newblock In \emph{Proceedings of Robotics: Science and Systems (RSS)}, 2023.

\bibitem[Collaboration et~al.(2024)Collaboration, O’Neill, Rehman, Maddukuri, Gupta, Padalkar, Lee, Pooley, Gupta, Mandlekar, Jain, et~al.]{o2024open}
Open X-Embodiment Collaboration, Abby O’Neill, Abdul Rehman, Abhiram Maddukuri, Abhishek Gupta, Abhishek Padalkar, Abraham Lee, Acorn Pooley, Agrim Gupta, Ajay Mandlekar, Ajinkya Jain, et~al.
\newblock Open x-embodiment: Robotic learning datasets and rt-x models.
\newblock In \emph{Proceedings of the IEEE International Conference on Robotics and Automation (ICRA)}, 2024.

\bibitem[Cui et~al.(2023)Cui, Wang, Shafiullah, and Pinto]{cuiplay}
Zichen~Jeff Cui, Yibin Wang, Nur Muhammad~Mahi Shafiullah, and Lerrel Pinto.
\newblock From play to policy: Conditional behavior generation from uncurated robot data.
\newblock In \emph{Proceedings of International Conference on Learning Representations (ICLR)}, 2023.

\bibitem[Dass et~al.(2023)Dass, Yapeter, Zhang, Zhang, Pertsch, Nikolaidis, and Lim]{dass2023jacoplay}
Shivin Dass, Jullian Yapeter, Jesse Zhang, Jiahui Zhang, Karl Pertsch, Stefanos Nikolaidis, and Joseph~J. Lim.
\newblock Clvr jaco play dataset, 2023.
\newblock URL \url{https://github.com/clvrai/clvr_jaco_play_dataset}.

\bibitem[Doshi et~al.(2024)Doshi, Walke, Mees, Dasari, and Levine]{doshi2024scaling}
Ria Doshi, Homer Walke, Oier Mees, Sudeep Dasari, and Sergey Levine.
\newblock Scaling cross-embodied learning: One policy for manipulation, navigation, locomotion and aviation.
\newblock In \emph{Proceedings of the Conference on Robot Learning (CoRL)}, 2024.

\bibitem[Ebert et~al.(2022)Ebert, Yang, Schmeckpeper, Bucher, Georgakis, Daniilidis, Finn, and Levine]{Ebert2021BridgeDB}
Frederik Ebert, Yanlai Yang, Karl Schmeckpeper, Bernadette Bucher, Georgios Georgakis, Kostas Daniilidis, Chelsea Finn, and Sergey Levine.
\newblock Bridge data: Boosting generalization of robotic skills with cross-domain datasets.
\newblock In \emph{Proceedings of Robotics: Science and Systems (RSS)}, 2022.

\bibitem[Fang et~al.(2023)Fang, Fang, Tang, Liu, Wang, Zhu, and Lu]{fang2023rh20t}
Hao-Shu Fang, Hongjie Fang, Zhenyu Tang, Jirong Liu, Junbo Wang, Haoyi Zhu, and Cewu Lu.
\newblock Rh20t: A comprehensive robotic dataset for learning diverse skills in one-shot.
\newblock In \emph{RSS 2023 Workshop on Learning for Task and Motion Planning}, 2023.

\bibitem[Fu et~al.(2024)Fu, Liu, Chen, Nie, and Xiong]{fu2024scene}
Rao Fu, Jingyu Liu, Xilun Chen, Yixin Nie, and Wenhan Xiong.
\newblock Scene-llm: Extending language model for 3d visual understanding and reasoning.
\newblock \emph{arXiv preprint arXiv:2403.11401}, 2024.

\bibitem[Gallistel(1990)]{gallistel1990organization}
Charles~R Gallistel.
\newblock \emph{The organization of learning.}
\newblock The MIT Press, 1990.

\bibitem[Haldar and Pinto(2023)]{haldar2023polytask}
Siddhant Haldar and Lerrel Pinto.
\newblock Polytask: Learning unified policies through behavior distillation.
\newblock \emph{arXiv preprint arXiv:2310.08573}, 2023.

\bibitem[Haldar et~al.(2024)Haldar, Peng, and Pinto]{haldar2024baku}
Siddhant Haldar, Zhuoran Peng, and Lerrel Pinto.
\newblock Baku: An efficient transformer for multi-task policy learning.
\newblock In \emph{Proceedings of the Conference on Neural Information Processing System (NeurIPS)}, 2024.

\bibitem[Heo et~al.(2023)Heo, Lee, Lee, and Lim]{heo2023furniturebench}
Minho Heo, Youngwoon Lee, Doohyun Lee, and Joseph~J Lim.
\newblock Furniturebench: Reproducible real-world benchmark for long-horizon complex manipulation.
\newblock In \emph{Proceedings of Robotics: Science and Systems (RSS)}, 2023.

\bibitem[Hong et~al.(2023)Hong, Zhen, Chen, Zheng, Du, Chen, and Gan]{hong20233d}
Yining Hong, Haoyu Zhen, Peihao Chen, Shuhong Zheng, Yilun Du, Zhenfang Chen, and Chuang Gan.
\newblock 3d-llm: Injecting the 3d world into large language models.
\newblock In \emph{Proceedings of the Conference on Neural Information Processing System (NeurIPS)}, 2023.

\bibitem[Huang et~al.(2024)Huang, Yong, Ma, Linghu, Li, Wang, Li, Zhu, Jia, and Huang]{huang2023embodied}
Jiangyong Huang, Silong Yong, Xiaojian Ma, Xiongkun Linghu, Puhao Li, Yan Wang, Qing Li, Song-Chun Zhu, Baoxiong Jia, and Siyuan Huang.
\newblock An embodied generalist agent in 3d world.
\newblock In \emph{Proceedings of the International Conference on Machine Learning (ICML)}, 2024.

\bibitem[Jang et~al.(2022)Jang, Irpan, Khansari, Kappler, Ebert, Lynch, Levine, and Finn]{jang2022bc}
Eric Jang, Alex Irpan, Mohi Khansari, Daniel Kappler, Frederik Ebert, Corey Lynch, Sergey Levine, and Chelsea Finn.
\newblock Bc-z: Zero-shot task generalization with robotic imitation learning.
\newblock In \emph{Proceedings of the Conference on Robot Learning (CoRL)}, 2022.

\bibitem[Kalashnikov et~al.(2018)Kalashnikov, Irpan, Pastor, Ibarz, Herzog, Jang, Quillen, Holly, Kalakrishnan, Vanhoucke, et~al.]{kalashnikov2018scalable}
Dmitry Kalashnikov, Alex Irpan, Peter Pastor, Julian Ibarz, Alexander Herzog, Eric Jang, Deirdre Quillen, Ethan Holly, Mrinal Kalakrishnan, Vincent Vanhoucke, et~al.
\newblock Qt-opt: Scalable deep reinforcement learning for vision-based robotic manipulation.
\newblock In \emph{Proceedings of the Conference on Robot Learning (CoRL)}, 2018.

\bibitem[Karamcheti et~al.(2024)Karamcheti, Nair, Balakrishna, Liang, Kollar, and Sadigh]{karamcheti2024prismatic}
Siddharth Karamcheti, Suraj Nair, Ashwin Balakrishna, Percy Liang, Thomas Kollar, and Dorsa Sadigh.
\newblock Prismatic vlms: Investigating the design space of visually-conditioned language models.
\newblock In \emph{Proceedings of the International Conference on Machine Learning (ICML)}, 2024.

\bibitem[Khazatsky et~al.(2024)Khazatsky, Pertsch, Nair, Balakrishna, Dasari, Karamcheti, Nasiriany, Srirama, Chen, Ellis, et~al.]{khazatsky2024droid}
Alexander Khazatsky, Karl Pertsch, Suraj Nair, Ashwin Balakrishna, Sudeep Dasari, Siddharth Karamcheti, Soroush Nasiriany, Mohan~Kumar Srirama, Lawrence~Yunliang Chen, Kirsty Ellis, et~al.
\newblock Droid: A large-scale in-the-wild robot manipulation dataset.
\newblock \emph{arXiv preprint arXiv:2403.12945}, 2024.

\bibitem[Kim et~al.(2024)Kim, Pertsch, Karamcheti, Xiao, Balakrishna, Nair, Rafailov, Foster, Lam, Sanketi, et~al.]{kim2024openvla}
Moo~Jin Kim, Karl Pertsch, Siddharth Karamcheti, Ted Xiao, Ashwin Balakrishna, Suraj Nair, Rafael Rafailov, Ethan Foster, Grace Lam, Pannag Sanketi, et~al.
\newblock Openvla: An open-source vision-language-action model.
\newblock \emph{arXiv preprint arXiv:2406.09246}, 2024.

\bibitem[Li et~al.(2024{\natexlab{a}})Li, Liang, Wang, Luo, Chen, Liao, Wei, Deng, Xu, Zhang, et~al.]{li2024cogact}
Qixiu Li, Yaobo Liang, Zeyu Wang, Lin Luo, Xi~Chen, Mozheng Liao, Fangyun Wei, Yu~Deng, Sicheng Xu, Yizhong Zhang, et~al.
\newblock Cogact: A foundational vision-language-action model for synergizing cognition and action in robotic manipulation.
\newblock \emph{arXiv preprint arXiv:2411.19650}, 2024{\natexlab{a}}.

\bibitem[Li et~al.(2024{\natexlab{b}})Li, Li, Liu, Wang, Liu, Kang, Ma, Kong, Zhang, and Liu]{li2023generalist}
Xinghang Li, Peiyan Li, Minghuan Liu, Dong Wang, Jirong Liu, Bingyi Kang, Xiao Ma, Tao Kong, Hanbo Zhang, and Huaping Liu.
\newblock Towards generalist robot policies: What matters in building vision-language-action models.
\newblock \emph{arXiv preprint arXiv:2412.14058}, 2024{\natexlab{b}}.

\bibitem[Li et~al.(2024{\natexlab{c}})Li, Li, Liu, Wang, Liu, Kang, Ma, Kong, Zhang, and Liu]{li2024towards}
Xinghang Li, Peiyan Li, Minghuan Liu, Dong Wang, Jirong Liu, Bingyi Kang, Xiao Ma, Tao Kong, Hanbo Zhang, and Huaping Liu.
\newblock Towards generalist robot policies: What matters in building vision-language-action models.
\newblock \emph{arXiv preprint arXiv:2412.14058}, 2024{\natexlab{c}}.

\bibitem[Li et~al.(2024{\natexlab{d}})Li, Liu, Zhang, Yu, Xu, Wu, Cheang, Jing, Zhang, Liu, et~al.]{livision}
Xinghang Li, Minghuan Liu, Hanbo Zhang, Cunjun Yu, Jie Xu, Hongtao Wu, Chilam Cheang, Ya~Jing, Weinan Zhang, Huaping Liu, et~al.
\newblock Vision-language foundation models as effective robot imitators.
\newblock In \emph{Proceedings of International Conference on Learning Representations (ICLR)}, 2024{\natexlab{d}}.

\bibitem[Li et~al.(2024{\natexlab{e}})Li, Hsu, Gu, Pertsch, Mees, Walke, Fu, Lunawat, Sieh, Kirmani, Levine, Wu, Finn, Su, Vuong, and Xiao]{li24simpler}
Xuanlin Li, Kyle Hsu, Jiayuan Gu, Karl Pertsch, Oier Mees, Homer~Rich Walke, Chuyuan Fu, Ishikaa Lunawat, Isabel Sieh, Sean Kirmani, Sergey Levine, Jiajun Wu, Chelsea Finn, Hao Su, Quan Vuong, and Ted Xiao.
\newblock Evaluating real-world robot manipulation policies in simulation.
\newblock In \emph{Proceedings of the Conference on Robot Learning (CoRL)}, 2024{\natexlab{e}}.

\bibitem[Liu et~al.(2023{\natexlab{a}})Liu, Zhu, Gao, Feng, Liu, Zhu, and Stone]{liu2023libero}
Bo~Liu, Yifeng Zhu, Chongkai Gao, Yihao Feng, Qiang Liu, Yuke Zhu, and Peter Stone.
\newblock Libero: Benchmarking knowledge transfer for lifelong robot learning.
\newblock \emph{arXiv preprint arXiv:2306.03310}, 2023{\natexlab{a}}.

\bibitem[Liu et~al.(2024{\natexlab{a}})Liu, Li, Wu, and Lee]{liu2024visual}
Haotian Liu, Chunyuan Li, Qingyang Wu, and Yong~Jae Lee.
\newblock Visual instruction tuning.
\newblock In \emph{Proceedings of the Conference on Neural Information Processing System (NeurIPS)}, 2024{\natexlab{a}}.

\bibitem[Liu et~al.(2023{\natexlab{b}})Liu, Nasiriany, Zhang, Bao, and Zhu]{liu2022robot}
Huihan Liu, Soroush Nasiriany, Lance Zhang, Zhiyao Bao, and Yuke Zhu.
\newblock Robot learning on the job: Human-in-the-loop autonomy and learning during deployment.
\newblock In \emph{Proceedings of Robotics: Science and Systems (RSS)}, 2023{\natexlab{b}}.

\bibitem[Liu et~al.(2024{\natexlab{b}})Liu, Wu, Li, Tan, Chen, Wang, Xu, Su, and Zhu]{liu2024rdt}
Songming Liu, Lingxuan Wu, Bangguo Li, Hengkai Tan, Huayu Chen, Zhengyi Wang, Ke~Xu, Hang Su, and Jun Zhu.
\newblock Rdt-1b: a diffusion foundation model for bimanual manipulation.
\newblock \emph{arXiv preprint arXiv:2410.07864}, 2024{\natexlab{b}}.

\bibitem[Logie(2014)]{logie2014visuo}
Robert~H Logie.
\newblock \emph{Visuo-spatial working memory}.
\newblock Psychology Press, 2014.

\bibitem[Luo et~al.(2024{\natexlab{a}})Luo, Xu, Geng, Feng, Fang, Tan, Schaal, and Levine]{luo2024multi}
Jianlan Luo, Charles Xu, Xinyang Geng, Gilbert Feng, Kuan Fang, Liam Tan, Stefan Schaal, and Sergey Levine.
\newblock Multi-stage cable routing through hierarchical imitation learning.
\newblock \emph{IEEE Transactions on Robotics}, 40:\penalty0 1476--1491, 2024{\natexlab{a}}.

\bibitem[Luo et~al.(2024{\natexlab{b}})Luo, Xu, Liu, Tan, Lin, Wu, Abbeel, and Levine]{luo2023fmb}
Jianlan Luo, Charles Xu, Fangchen Liu, Liam Tan, Zipeng Lin, Jeffrey Wu, Pieter Abbeel, and Sergey Levine.
\newblock Fmb: a functional manipulation benchmark for generalizable robotic learning.
\newblock \emph{The International Journal of Robotics Research}, 2024{\natexlab{b}}.

\bibitem[Lynch et~al.(2023)Lynch, Wahid, Tompson, Ding, Betker, Baruch, Armstrong, and Florence]{lynch2023interactive}
Corey Lynch, Ayzaan Wahid, Jonathan Tompson, Tianli Ding, James Betker, Robert Baruch, Travis Armstrong, and Pete Florence.
\newblock Interactive language: Talking to robots in real time.
\newblock \emph{IEEE Robotics and Automation Letters}, 2023.

\bibitem[Mandlekar et~al.(2018)Mandlekar, Zhu, Garg, Booher, Spero, Tung, Gao, Emmons, Gupta, Orbay, et~al.]{mandlekar2018roboturk}
Ajay Mandlekar, Yuke Zhu, Animesh Garg, Jonathan Booher, Max Spero, Albert Tung, Julian Gao, John Emmons, Anchit Gupta, Emre Orbay, et~al.
\newblock Roboturk: A crowdsourcing platform for robotic skill learning through imitation.
\newblock In \emph{Proceedings of the Conference on Robot Learning (CoRL)}, 2018.

\bibitem[Mees et~al.(2023)Mees, Borja-Diaz, and Burgard]{mees2023grounding}
Oier Mees, Jessica Borja-Diaz, and Wolfram Burgard.
\newblock Grounding language with visual affordances over unstructured data.
\newblock In \emph{Proceedings of the IEEE International Conference on Robotics and Automation (ICRA)}, 2023.

\bibitem[Mendonca et~al.(2023)Mendonca, Bahl, and Pathak]{mendoncastructured}
Russell Mendonca, Shikhar Bahl, and Deepak Pathak.
\newblock Structured world models from human videos.
\newblock In \emph{Proceedings of the Conference on Robot Learning (CoRL)}, 2023.

\bibitem[Nasiriany et~al.(2023)Nasiriany, Gao, Mandlekar, and Zhu]{nasiriany2023learning}
Soroush Nasiriany, Tian Gao, Ajay Mandlekar, and Yuke Zhu.
\newblock Learning and retrieval from prior data for skill-based imitation learning.
\newblock In \emph{Proceedings of the Conference on Robot Learning (CoRL)}, 2023.

\bibitem[{Octo Model Team} et~al.(2024){Octo Model Team}, Ghosh, Walke, Pertsch, Black, Mees, Dasari, Hejna, Xu, Luo, Kreiman, Tan, Chen, Sanketi, Vuong, Xiao, Sadigh, Finn, and Levine]{team2024octo}
{Octo Model Team}, Dibya Ghosh, Homer Walke, Karl Pertsch, Kevin Black, Oier Mees, Sudeep Dasari, Joey Hejna, Charles Xu, Jianlan Luo, Tobias Kreiman, {You Liang} Tan, Lawrence~Yunliang Chen, Pannag Sanketi, Quan Vuong, Ted Xiao, Dorsa Sadigh, Chelsea Finn, and Sergey Levine.
\newblock Octo: An open-source generalist robot policy.
\newblock In \emph{Proceedings of Robotics: Science and Systems (RSS)}, 2024.

\bibitem[Parisotto et~al.(2016)Parisotto, Ba, and Salakhutdinov]{parisotto2015actor}
Emilio Parisotto, Jimmy~Lei Ba, and Ruslan Salakhutdinov.
\newblock Actor-mimic: Deep multitask and transfer reinforcement learning.
\newblock In \emph{Proceedings of International Conference on Learning Representations (ICLR)}, 2016.

\bibitem[Peng et~al.(2023)Peng, Wang, Dong, Hao, Huang, Ma, and Wei]{peng2023kosmos}
Zhiliang Peng, Wenhui Wang, Li~Dong, Yaru Hao, Shaohan Huang, Shuming Ma, and Furu Wei.
\newblock Kosmos-2: Grounding multimodal large language models to the world.
\newblock \emph{arXiv preprint arXiv:2306.14824}, 2023.

\bibitem[Pertsch et~al.(2025)Pertsch, Stachowicz, Ichter, Driess, Nair, Vuong, Mees, Finn, and Levine]{pertsch2025fast}
Karl Pertsch, Kyle Stachowicz, Brian Ichter, Danny Driess, Suraj Nair, Quan Vuong, Oier Mees, Chelsea Finn, and Sergey Levine.
\newblock Fast: Efficient action tokenization for vision-language-action models.
\newblock \emph{arXiv preprint arXiv:2501.09747}, 2025.

\bibitem[Piaget(2013)]{piaget2013child}
Jean Piaget.
\newblock \emph{Child's Conception of Space: Selected Works vol 4}.
\newblock Routledge, 2013.

\bibitem[Qu et~al.(2024)Qu, Chen, Zhang, Gao, Li, Zhao, Wang, and Li]{qu2024livescene}
Delin Qu, Qizhi Chen, Pingrui Zhang, Xianqiang Gao, Junzhe Li, Bin Zhao, Dong Wang, and Xuelong Li.
\newblock Livescene: Language embedding interactive radiance fields for physical scene rendering and control.
\newblock \emph{arXiv preprint arXiv:2406.16038}, 2024.

\bibitem[Quere et~al.(2020)Quere, Hagengruber, Iskandar, Bustamante, Leidner, Stulp, and Vogel]{quere2020shared}
Gabriel Quere, Annette Hagengruber, Maged Iskandar, Samuel Bustamante, Daniel Leidner, Freek Stulp, and J{\"o}rn Vogel.
\newblock Shared control templates for assistive robotics.
\newblock In \emph{Proceedings of the IEEE International Conference on Robotics and Automation (ICRA)}, 2020.

\bibitem[Radford et~al.(2021)Radford, Kim, Hallacy, Ramesh, Goh, Agarwal, Sastry, Askell, Mishkin, Clark, et~al.]{radford2021learning}
Alec Radford, Jong~Wook Kim, Chris Hallacy, Aditya Ramesh, Gabriel Goh, Sandhini Agarwal, Girish Sastry, Amanda Askell, Pamela Mishkin, Jack Clark, et~al.
\newblock Learning transferable visual models from natural language supervision.
\newblock In \emph{Proceedings of the International Conference on Machine Learning (ICML)}, 2021.

\bibitem[Rosete-Beas et~al.(2022)Rosete-Beas, Mees, Kalweit, Boedecker, and Burgard]{rosete2023latent}
Erick Rosete-Beas, Oier Mees, Gabriel Kalweit, Joschka Boedecker, and Wolfram Burgard.
\newblock Latent plans for task-agnostic offline reinforcement learning.
\newblock In \emph{Proceedings of the Conference on Robot Learning (CoRL)}, 2022.

\bibitem[Rusu et~al.(2016)Rusu, Colmenarejo, Gulcehre, Desjardins, Kirkpatrick, Pascanu, Mnih, Kavukcuoglu, and Hadsell]{rusu2015policy}
Andrei~A Rusu, Sergio~Gomez Colmenarejo, Caglar Gulcehre, Guillaume Desjardins, James Kirkpatrick, Razvan Pascanu, Volodymyr Mnih, Koray Kavukcuoglu, and Raia Hadsell.
\newblock Policy distillation.
\newblock In \emph{Proceedings of International Conference on Learning Representations (ICLR)}, 2016.

\bibitem[Saxena et~al.(2023)Saxena, Sharma, and Kroemer]{saxena2023multi}
Saumya Saxena, Mohit Sharma, and Oliver Kroemer.
\newblock Multi-resolution sensing for real-time control with vision-language models.
\newblock In \emph{Proceedings of the Conference on Robot Learning (CoRL)}, 2023.

\bibitem[Shafiullah et~al.(2023)Shafiullah, Rai, Etukuru, Liu, Misra, Chintala, and Pinto]{shafiullah2023bringing}
Nur Muhammad~Mahi Shafiullah, Anant Rai, Haritheja Etukuru, Yiqian Liu, Ishan Misra, Soumith Chintala, and Lerrel Pinto.
\newblock On bringing robots home.
\newblock \emph{arXiv preprint arXiv:2311.16098}, 2023.

\bibitem[Shah et~al.(2023)Shah, Mart{\'\i}n-Mart{\'\i}n, and Zhu]{shahmutex}
Rutav Shah, Roberto Mart{\'\i}n-Mart{\'\i}n, and Yuke Zhu.
\newblock Mutex: Learning unified policies from multimodal task specifications.
\newblock In \emph{Proceedings of the Conference on Robot Learning (CoRL)}, 2023.

\bibitem[Shridhar et~al.(2022)Shridhar, Manuelli, and Fox]{shridhar2023perceiver}
Mohit Shridhar, Lucas Manuelli, and Dieter Fox.
\newblock Perceiver-actor: A multi-task transformer for robotic manipulation.
\newblock In \emph{Proceedings of the Conference on Robot Learning (CoRL)}, 2022.

\bibitem[Steiner et~al.(2024)Steiner, Pinto, Tschannen, Keysers, Wang, Bitton, Gritsenko, Minderer, Sherbondy, Long, et~al.]{steiner2024paligemma}
Andreas Steiner, Andr{\'e}~Susano Pinto, Michael Tschannen, Daniel Keysers, Xiao Wang, Yonatan Bitton, Alexey Gritsenko, Matthias Minderer, Anthony Sherbondy, Shangbang Long, et~al.
\newblock Paligemma 2: A family of versatile vlms for transfer.
\newblock \emph{arXiv preprint arXiv:2412.03555}, 2024.

\bibitem[Tolman(1948)]{tolman1948cognitive}
Edward~C Tolman.
\newblock Cognitive maps in rats and men.
\newblock \emph{Psychological review}, 55\penalty0 (4):\penalty0 189, 1948.

\bibitem[Walke et~al.(2023)Walke, Black, Lee, Kim, Du, Zheng, Zhao, Hansen-Estruch, Vuong, He, Myers, Fang, Finn, and Levine]{walke2023bridgedata}
Homer Walke, Kevin Black, Abraham Lee, Moo~Jin Kim, Max Du, Chongyi Zheng, Tony Zhao, Philippe Hansen-Estruch, Quan Vuong, Andre He, Vivek Myers, Kuan Fang, Chelsea Finn, and Sergey Levine.
\newblock Bridgedata v2: A dataset for robot learning at scale.
\newblock In \emph{Proceedings of the Conference on Robot Learning (CoRL)}, 2023.

\bibitem[Wang et~al.(2024)Wang, Chen, Zhao, and He]{wang2024scaling}
Lirui Wang, Xinlei Chen, Jialiang Zhao, and Kaiming He.
\newblock Scaling proprioceptive-visual learning with heterogeneous pre-trained transformers.
\newblock In \emph{Proceedings of the Conference on Neural Information Processing System (NeurIPS)}, 2024.

\bibitem[Yan et~al.(2023)Yan, Wu, and Wang]{ucsdkitchens}
Ge~Yan, Kris Wu, and Xiaolong Wang.
\newblock ucsd kitchens dataset.
\newblock \url{https://github.com/geyan21/rlds_dataset_builder/tree/main/ucsd_kitchens}, 2023.

\bibitem[Yang et~al.(2024)Yang, Yang, Gupta, Han, Fei-Fei, and Xie]{yang2024thinking}
Jihan Yang, Shusheng Yang, Anjali~W Gupta, Rilyn Han, Li~Fei-Fei, and Saining Xie.
\newblock Thinking in space: How multimodal large language models see, remember, and recall spaces.
\newblock \emph{arXiv preprint arXiv:2412.14171}, 2024.

\bibitem[Zhai et~al.(2023)Zhai, Mustafa, Kolesnikov, and Beyer]{zhai2023sigmoid}
Xiaohua Zhai, Basil Mustafa, Alexander Kolesnikov, and Lucas Beyer.
\newblock Sigmoid loss for language image pre-training.
\newblock In \emph{Proceedings of the IEEE/CVF International Conference on Computer Vision (ICCV)}, 2023.

\bibitem[Zhen et~al.(2024)Zhen, Qiu, Chen, Yang, Yan, Du, Hong, and Gan]{zhen20243d}
Haoyu Zhen, Xiaowen Qiu, Peihao Chen, Jincheng Yang, Xin Yan, Yilun Du, Yining Hong, and Chuang Gan.
\newblock 3d-vla: A 3d vision-language-action generative world model.
\newblock In \emph{Proceedings of the International Conference on Machine Learning (ICML)}, 2024.

\bibitem[Zheng et~al.(2025)Zheng, Li, Liu, Zheng, Wang, Ou, Liu, Liu, Zhang, and Zhan]{zheng2025universal}
Jinliang Zheng, Jianxiong Li, Dongxiu Liu, Yinan Zheng, Zhihao Wang, Zhonghong Ou, Yu~Liu, Jingjing Liu, Ya-Qin Zhang, and Xianyuan Zhan.
\newblock Universal actions for enhanced embodied foundation models.
\newblock \emph{arXiv preprint arXiv:2501.10105}, 2025.

\bibitem[Zheng et~al.(2024)Zheng, Liang, Huang, Gao, Daum{\'e}~III, Kolobov, Huang, and Yang]{zheng2024tracevla}
Ruijie Zheng, Yongyuan Liang, Shuaiyi Huang, Jianfeng Gao, Hal Daum{\'e}~III, Andrey Kolobov, Furong Huang, and Jianwei Yang.
\newblock Tracevla: Visual trace prompting enhances spatial-temporal awareness for generalist robotic policies.
\newblock \emph{arXiv preprint arXiv:2412.10345}, 2024.

\bibitem[Zhou et~al.(2023)Zhou, Dean, Srirama, Rajeswaran, Pari, Hatch, Jain, Yu, Abbeel, Pinto, et~al.]{zhou2023train}
Gaoyue Zhou, Victoria Dean, Mohan~Kumar Srirama, Aravind Rajeswaran, Jyothish Pari, Kyle Hatch, Aryan Jain, Tianhe Yu, Pieter Abbeel, Lerrel Pinto, et~al.
\newblock Train offline, test online: A real robot learning benchmark.
\newblock In \emph{Proceedings of the IEEE International Conference on Robotics and Automation (ICRA)}, 2023.

\bibitem[Zhu et~al.(2024)Zhu, Wang, Zhang, Pang, and Liu]{zhu2024llava}
Chenming Zhu, Tai Wang, Wenwei Zhang, Jiangmiao Pang, and Xihui Liu.
\newblock Llava-3d: A simple yet effective pathway to empowering lmms with 3d-awareness.
\newblock \emph{arXiv preprint arXiv:2409.18125}, 2024.

\bibitem[Zhu et~al.(2023{\natexlab{a}})Zhu, Tian, Xu, Huo, Zhan, Tomizuka, and Ding]{zhu2023fanuc}
Xinghao Zhu, Ran Tian, Chenfeng Xu, Mingxiao Huo, Wei Zhan, Masayoshi Tomizuka, and Mingyu Ding.
\newblock Fanuc manipulation: A dataset for learning-based manipulation with fanuc mate 200id robot.
\newblock \url{https://sites.google.com/berkeley.edu/fanuc-manipulation}, 2023{\natexlab{a}}.

\bibitem[Zhu et~al.(2022)Zhu, Stone, and Zhu]{zhu2022bottom}
Yifeng Zhu, Peter Stone, and Yuke Zhu.
\newblock Bottom-up skill discovery from unsegmented demonstrations for long-horizon robot manipulation.
\newblock \emph{IEEE Robotics and Automation Letters}, 7\penalty0 (2):\penalty0 4126--4133, 2022.

\bibitem[Zhu et~al.(2023{\natexlab{b}})Zhu, Jiang, Stone, and Zhu]{zhu2023learning}
Yifeng Zhu, Zhenyu Jiang, Peter Stone, and Yuke Zhu.
\newblock Learning generalizable manipulation policies with object-centric 3d representations.
\newblock In \emph{Proceedings of the Conference on Robot Learning (CoRL)}, 2023{\natexlab{b}}.

\bibitem[Zhu et~al.(2023{\natexlab{c}})Zhu, Joshi, Stone, and Zhu]{zhu2023viola}
Yifeng Zhu, Abhishek Joshi, Peter Stone, and Yuke Zhu.
\newblock Viola: Imitation learning for vision-based manipulation with object proposal priors.
\newblock In \emph{Proceedings of the Conference on Robot Learning (CoRL)}, 2023{\natexlab{c}}.

\end{thebibliography}

\clearpage

\maketitlesupplementary
\appendix
\setcounter{page}{1}
\subsection{Dataset Mixture Details}
\label{supp:data_mixture} 
\begin{figure}[htbp]
   \vspace{-2ex}
   \begin{center}
      \includegraphics[width=0.9\linewidth]{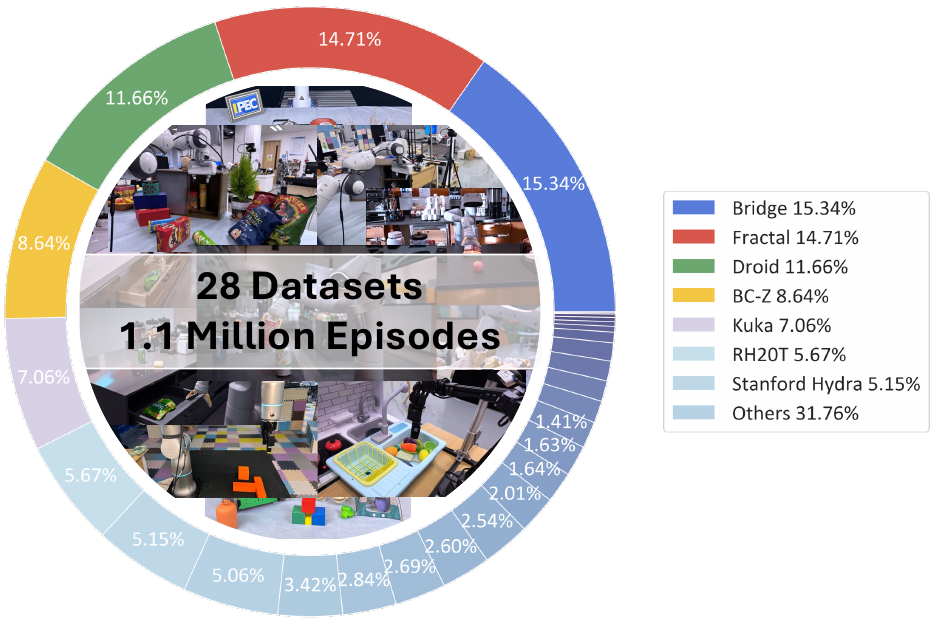}
   \end{center}
   \vspace{-1.5ex}
   \caption{SpatialVLA Training Dataset Mixture Visualization.}
   \label{fig:dataset}
   \vspace{-2.5ex}
\end{figure}

~\cref{fig:dataset} illustrates the dataset mixture for
\ours. The training dataset is primarily composed of: Bridge (15.34\%), Fractal
(14.71\%), Droid (11.66\%), BC-Z (8.64\%), Kuka (7.06\%), RH20T (5.67\%), Stanford
Hydra (5.15\%) and Language Table (5.06\%). We extended OpenVLA's data mixing by
adding RH20T~\citep{fang2023rh20t} and training/testing individual datasets on a
real robot evaluation, down-weighting Kuka, Toto, Berkeley Fanuc Manipulation, and
FMB Dataset~\citep{luo2023fmb}. Notably, we found that an excessive proportion
of the FMB Dataset led to a significant rightward bias in the robot, and the Kuka dataset
lacked clear prompts.

\begin{table}[htbp]
    \centering
    \caption{\textbf{\ours Dataset Mixtures Details.}}
    \begin{tabular}{l|r|c|c}
        \toprule
        Dataset                                                       & Weight  & trajectory & sample   \\
        \midrule Bridge~\citep{walke2023bridgedata,Ebert2021BridgeDB} & 15.34\% & 60064      & 2135463  \\
        Fractal~\citep{brohan2022rt}                                  & 14.71\% & 87212      & 3786400  \\
        Droid~\citep{khazatsky2024droid}                              & 11.66\% & 92233      & 27044326 \\
        BC-Z~\citep{jang2022bc}                                       & 8.64\%  & 43264      & 6015535  \\
        Kuka~\citep{kalashnikov2018scalable}                          & 7.06\%  & 209880     & 2455879  \\
        RH20T~\citep{fang2023rh20t}                                   & 5.67\%  & 104392     & 52644433 \\
        Stanford Hydra~\citep{belkhale2023hydra}                      & 5.15\%  & 570        & 358234   \\
        Language Table~\citep{lynch2023interactive}                   & 5.06\%  & 442226     & 7045476  \\
        Taco Play~\citep{rosete2023latent,mees2023grounding}          & 3.42\%  & 3603       & 237798   \\
        Furniture Bench~\citep{heo2023furniturebench}                 & 2.84\%  & 5100       & 3948057  \\
        Roboturk~\citep{mandlekar2018roboturk}                        & 2.69\%  & 1995       & 187507   \\
        Utaustin Mutex~\citep{shahmutex}                              & 2.60\%  & 1500       & 361883   \\
        Austin Sailor~\citep{nasiriany2023learning}                   & 2.54\%  & 240        & 353094   \\
        Austin Sirius~\citep{liu2022robot}                            & 2.01\%  & 559        & 279939   \\
        DobbE~\citep{shafiullah2023bringing}                          & 1.64\%  & 5208       & 1139911  \\
        FMB Dataset~\citep{luo2023fmb}                                & 1.63\%  & 8612       & 1137459  \\
        Berkeley Autolab UR5~\citep{BerkeleyUR5Website}               & 1.41\%  & 1000       & 97939    \\
        Toto~\citep{zhou2023train}                                    & 1.17\%  & 1003       & 325699   \\
        Viola~\citep{zhu2023viola}                                    & 1.10\%  & 150        & 76324    \\
        IAMLab CMU Pickup Insert~\citep{saxena2023multi}              & 1.05\%  & 631        & 146241   \\
        NYU Franka~\citep{cuiplay}                                    & 0.97\%  & 456        & 44875    \\
        Jaco Play~\citep{dass2023jacoplay}                            & 0.56\%  & 1085       & 77965    \\
        Berkeley Cable Routing~\citep{luo2024multi}                   & 0.30\%  & 1647       & 42328    \\
        Austin Buds~\citep{zhu2022bottom}                             & 0.25\%  & 50         & 34112    \\
        Berkeley Fanuc Manipulation~\citep{zhu2023fanuc}              & 0.22\%  & 415        & 62613    \\
        CMU Stretch~\citep{mendoncastructured}                        & 0.18\%  & 135        & 25016    \\
        DLR EDAN Shared Control~\citep{quere2020shared}               & 0.06\%  & 104        & 8928     \\
        UCSD Kitchen~\citep{ucsdkitchens}                             & 0.06\%  & 150        & 3970     \\
        \bottomrule
    \end{tabular}
    \label{tab:dataset_mixture}
\end{table}

\begin{algorithm*}
   [t]
   \caption{Python pseudocode for \ours action encoding and decoding.}
   \label{alg:tracevla_inference}
   \definecolor{codecomment}{rgb}{0.25,0.5,0.5}
   \definecolor{codekeyword}{rgb}{0.6,0.25,0.6}
   \lstset{
     basicstyle=\fontsize{7.2pt}{7.2pt}\ttfamily\bfseries,
     commentstyle=\fontsize{7.2pt}{7.2pt}\color{codecomment},
     keywordstyle=\fontsize{7.2pt}{7.2pt}\color{codekeyword},
     backgroundcolor=\color{white},  
   }
   \begin{lstlisting}[language=python]
# N: Number of Grid Intervals (e.g., 8)
# R: Range for Grids (e.g., [0, pi])
# P: Probability of Each Grids
# G: Adaptive Action Grids with Embedding of Size E
# cdf: Cumulative Distribution Function
# ppf: Percent Point Function

# create adptive action grids from gaussian distributions
for gaussians, grid_params in GS(theta, phi, r, roll, pitch, yaw):
   for (mu, sigma), (R, N)  in gaussians, grid_params:
         P = linspace(cdf(R, mu, sigma), cdf(R, mu, sigma), N + 1)
         G.x = ppf(P, mu, sigma) # coordinates
         G.fea = Embedding(N, E) # features

G.add_gripper() # add gripper 2 grids
# linearization 3d grids to share parameters with llm embedding
# trans: [N_theta * N_phi * N_r], rot: [N_roll * N_pitch * N_yaw]
# gripper: [N_gripper]
G.linearization()

# T: Number of Timesteps
# D: Dataset
for t in range(0, T): 
   # if encode
   a = D(t) # normalized action [theta, phi, r, roll, pitch, yaw, gripper]
      
   # digitize continuous actions to 3d grids
   d_theta, d_phi, d_r = digitize(G, theta, phi, r) # trans
   d_roll, d_pitch, d_yaw = digitize(G, roll, pitch, yaw) # rot
   
   # linearization
   id_trans = linearize(d_theta, d_phi, d_r)
   id_rot = linearize(d_roll, d_pitch, d_yaw)
   id_gripper = 1 if gripper > 0.5 else 0 # gripper
   token_trans, token_rot, token_gripper = G.fea(id_trans, id_rot, id_gripper)
   
   # if decode
   (id_trans, id_rot, id_gripper) = SpatialVLA([image], prompt) # predict 3 action token id
   d_theta, d_phi, d_r = gridification(G, id_trans)
   d_roll, d_pitch, d_yaw = gridification(G, id_rot)
   gripper = id_gripper
   a = unnomalize(d_theta, d_phi, d_r, d_roll, d_pitch, d_yaw, gripper)
\end{lstlisting}
   \label{alg:encode_decode}
\end{algorithm*}
\begin{table*}
    [thb!]
    \centering
    \caption{\textbf{Model Architecture of \ours}.}
    \resizebox{1.8\columnwidth}{!}{
    \begin{tabular}{c|c|c|cc}
        \hline
        Module                                           & Layer Type                 & Layer Num & Input Shape  & Output Shape          \\
        \hline
        \multicolumn{5}{c}{Siglip Vision Tower}           \\
        \hline
        \multirow{3}{*}{SiglipVisionTransformer}         & SiglipVisionEmbeddings     & 1         & (B, 3, H, W) & (B, 1152, H/14, W/14) \\
                                                         & SiglipEncoderLayer         & 27        & (B, N, 1152) & (B, N, 1152)          \\
                                                         & LayerNorm                  & 1         & (B, N, 1152) & (B, N, 1152)          \\
        \hline
        \multicolumn{5}{c}{SpatialVLAMultiModalProjector} \\
        \hline
        Linear                                           & Linear                     & 1         & (B, N, 1152) & (B, N, 2304)          \\
        \hline
        \multicolumn{5}{c}{Gemma2Model}                   \\
        \hline
        Embed Tokens                                     & Embedding                  & 1         & (B, T)       & (B, T, 2304)          \\
        Layers                                           & Gemma2DecoderLayer         & 26        & (B, S, 2304) & (B, S, 2304)          \\
        Norm                                             & Gemma2RMSNorm              & 1         & (B, S, 2304) & (B, S, 2304)          \\
        \hline
        \multicolumn{5}{c}{Spatial Embedding}             \\
        \hline
        Spatial Embedding                                & Embedding                  & 1         & (B, T)       & (B, T, 2304)          \\
        \hline
        \multicolumn{5}{c}{Vision Zoe Model}              \\
        \hline
        Vision Zoe Model                                 & ZoeDepthForDepthEstimation & 1         & (B, 3, H, W) & (B, 1, H, W)          \\
        \hline
        \multicolumn{5}{c}{Ego3DPositionEmbeddingMLP}     \\
        \hline
        \multirow{4}{*}{position\_embedding\_head}       & Linear                     & 1         & (B, 204)     & (B, 1152)             \\
                                                         & LayerNorm                  & 1         & (B, 1152)    & (B, 1152)             \\
                                                         & ReLU                       & 1         & (B, 1152)    & (B, 1152)             \\
                                                         & Linear                     & 1         & (B, 1152)    & (B, 1152)             \\
        \hline
    \end{tabular}
    } \label{tab:model_arch}
\end{table*}

\subsection{Model Architecture Details}
\label{supp:model_arch}

\noindent
\textbf{Model Architecture.} \cref{tab:model_arch} provides a detailed illustration
of the model components and input\/output of \ours. \ours employs Paligemma2 as
the backbone, consisting of Siglip Vision Tower, SpatialVLA MultiModal Projector,
Gemma2Model, Vision Zoe Model, Ego3D Position Embedding MLP, and Spatial
Embedding. The input RGB image of shape 224$\times$224 is separately fed into Siglip
Vision Tower and Vision Zoe Model to extract visual features and predict depth,
respectively. Subsequently, we back-project the depth into the point cloud, compute
the average high-frequency point cloud encoding per sub-patch, and project it
onto the Siglip image patch dimension with Ego3D Position Embedding MLP, followed
by addition.

\noindent
\textbf{Action Grids Encoding and Decoding.} We encode the action space into
adaptive action grids with Gaussian distributions and decode the action tokens
from the grids. ~\cref{alg:encode_decode} outlines the construction of \ours's
action grid and the action encoding-decoding pipeline. We discretize $\theta$,
$\phi$, and $r$ into 16, 32, and 8 bins, respectively, within their corresponding
ranges of $[0, \pi]$, $[-\pi, \pi]$, and $[0, \sqrt{3}]$. Similarly, roll, pitch,
and yaw are each divided into 16 bins within the range of $[-1, -1]$. Then, we calculate
the cumulative probability density (CDF) and compute the percent point function (PPF)
to achieve an exact grid discretization. Notably, we linearize the action spatial
grid into $V=8194$ spatial action token embeddings, sharing parameters with the LLM embedding, to enable seamless
model training. During encoding, we digitize rotational and translational actions
and compute linear grid indices, which are then queried to retrieve features for
training. Conversely, in decoding, we gridify the inferred indices into 3D positions
and unnormalize them into continuous actions.

\subsection{More Discussion}
\label{supp:more_discussion}

\begin{figure}[htbp]
    \vspace{-2ex}
    \begin{center}
        \includegraphics[trim=0.4ex 0 0 0, clip, width=1.0\linewidth]{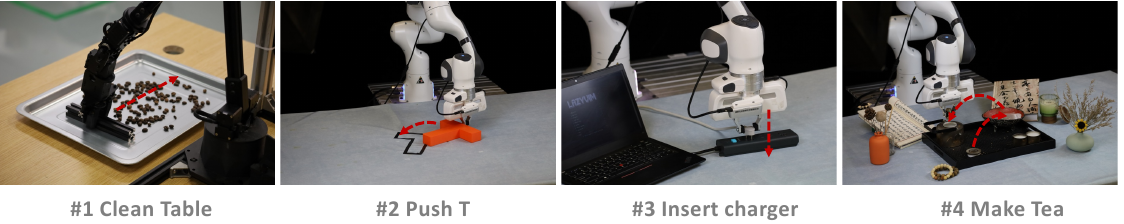}
    \end{center}
    \vspace{-2ex}
    \caption{Fine-grained manipulation beyond pick-and-place.}
    \label{fig:complex_task}
    \vspace{-2ex}
\end{figure}

\boldparagraph{Q1. \textcolor{Q}{\textit{Advantage and limitation of spatial discretization?}}}
\noindent $\triangleright$Autoregressive paradigm with discretized action token has been widely adopted in existing VLAs~\citep{brohan2023rt,kim2024openvla,li2024cogact}, demonstrating significant advantages in large-scale training, seamless LLM integration, and instruction-following capabilities. The proposed Adaptive Action Grids adaptively align heterogeneous robotic actions to 3D space (\cref{fig:pipeline}), delivering notable strengths in efficient action representation (\cref{fig:bin}), fast token decoding (\cref{fig:teaser}), and flexible post-training transfer (\cref{sec:training}). 
$\triangleright$\textit{A key concern is whether discretization limits precision.} As shown in \cref{tab:rebuttal_reso}, experiments across four grid resolutions demonstrate that performance improves from 1026 to 8194, but thereafter exhibits a gradual plateau or even a decline, beyond 6166. Notably, our method outperforms the 8196 uniform partitions \#1 $\mathcal{U}8196$ with only half partitions \#3 reso 4610, which is attributed to our \textit{adaptive grid partitioning strategy based on Gaussian probability density, allowing the model to concentrate on high-probability, fine-grained action spaces}, as depicted in~\cref{fig:bin}. Moreover, additional experiments in~\cref{fig:complex_task,tab:complex_task_real} show that \ours with discrete action space can handle fine-grained manipulation effectively. For the simplerenv~\cite{li24simpler} evaluation in~\cref{tab:simplerenv_bridge}, \ours also matches or outperforms the latest SOTA model $\pi_0$. $\triangleright$ However, \textit{higher-resolution or higher-dimensional action space introduces additional parameters to the 128k vocabulary, necessitating a trade-off between precision and efficiency}. The model may incur parameter overhead for humanoid robots with higher degrees of freedom. A flexible solution is to share action grids among embodiments like dexterous hands, bimanual arms, and legs, to be explored in future work.
\begin{table}[htbp]
    \centering
    \setlength{\tabcolsep}{2pt}
    \caption{Action discretization ablations on simplerenv~\cite{li24simpler}.}
    \vspace{-2ex}
    \resizebox{1\columnwidth}{!}{
    \begin{tabular}{l|cc|cc|cc|cc}
        \hline
        \#setting             & \multicolumn{2}{c|}{pick coke can} & \multicolumn{2}{c|}{move near} & \multicolumn{2}{c|}{put carrot} & \multicolumn{2}{c}{put eggplant} \\
        \hline
                              & aggregation                        & matching                       & aggregation                              & matching                                  & partial & success & partial & success \\
        $[1]$. $\mathcal{U}$8196 & 77.9\%                             & 28.0\%                         & 64.2\%                                   & 55.0\%                                    & 45.8\%  & 12.5\%  &  79.2\% & 54.2\%  \\
        $[2]$. reso 1026      & 74.4\%                             & 67.3\%                         & 59.1\%                                   & 54.2\%                                    & 45.8\%  & 25.0\%  & 66.7\%  & 54.2\%  \\
        $[3]$. reso 4610      & 76.7\%                             & 68.0\%                         & 69.8\%                                   & 79.2\%                                    & 41.7\%  & 33.3\%  & 83.3\%  & 75.0\%  \\
        $[4]$. reso 6166      & 80.9\%                             & 74.0\%                         & 74.0\%                                   & 79.2\%                                    & 41.7\%  & 33.3\%  & 95.8\%  & 87.5\%  \\
        $[5]$. reso 8194      & 81.6\%                             & 70.7\%                         & 79.2\%                                   & 85.4\%                                    & 41.7\%  & 33.3\%  & 91.7\%  & 87.5\%  \\
        \hline
    \end{tabular}
        } \label{tab:rebuttal_reso}
    \vspace{-2ex}
\end{table}

\boldparagraph{Q2. \textcolor{Q}{\textit{Handle more complex task beyond pick-and-place?}}}
\noindent
$\triangleright$ To address reviewers' concerns on generalizability and fine-grained tasks, we evaluate \ours with four new dexterity tasks in~\cref{fig:complex_task,tab:complex_task_real}. \ours performs well even in complex fine-grained manipulation, significantly outperforming OpenVLA~\citep{o2024open} (72.7\% vs 54.5\%). As discussed earlier, this shows the action grid resolution is not a critical limiting factor, benefiting from its probability-adaptive grids.

\begin{table}[t]
    \centering
    \setlength{\tabcolsep}{2pt}
    \caption{Fine-grained manipulation and depth ablations.}
    \vspace{-2ex}
    \resizebox{1\columnwidth}{!}{
    \begin{tabular}{l|c|c|c|c|c}
        \hline
        \#setting ($\mathbf{D}$:depth)& clean table & push-T & insert charger & make tea & avg \\
        \hline
        OpenVLA~\citep{kim2024openvla}                   & 72.7\%     & 72.7\%       & 18.2\%          & 54.5\%             & 54.5\%  \\
        \ours w/o $\mathbf{D}$    & 63.6\%     & 54.5\%       & 9.1\%          & 54.5\%             & 45.4\%  \\
        \ours sensor $\mathbf{D}$ & 90.9\%     & 81.8\%       & 27.3\%          & 81.8\%             & 70.5\%  \\
        \ours zoe $\mathbf{D}$    & 90.9\%     & 90.9\%       & 27.3\%          & 81.8\%             & 72.7\%  \\
        \hline
    \end{tabular}
    } 
    \label{tab:complex_task_real}
    \vspace{-4ex}
\end{table}

\boldparagraph{Q3. \textcolor{Q}{\textit{Complexity and noise analysis of ZoeDepth model?}}}
\noindent
$\triangleright$ We must emphasize that ZoeDepth is used because depth is unavailable for all 28 datasets in OXE~\cite{o2024open} and RH20T~\cite{fang2023rh20t}. In pretraining, we utilize ZoeDepth to predict depth and ensure feasibility. In contrast, downstream fine-tuning allows for either ZoeDepth or sensor-captured depth in~\cref{tab:complex_task_real}. Visualization results show that \textit{sensor depth is noisy and disruptive, while ZoeDepth provides smoother inputs}. Our model with ZoeDepth achieves 72.7\% ACC, surpassing models with sensor depth (70.5\%) and without depth (45.4\%). This confirms the significant benefits of Ego3D PE with ZoeDepth, which \textit{captures relative spatial layouts, rendering precise scale unnecessary}. Notably, ZoeDepth remains fixed during training, occupying only 8.6\% of parameters and 0.06s per action time cost, making it an efficient solution. Additionally, Layer Norm and zero weight initialization of Ego3D PE help mitigate noise issues.
\begin{figure}[htbp]
    \vspace{-2ex}
    \begin{center}
        \includegraphics[trim=0.4ex 0 0 0, clip, width=1\linewidth]{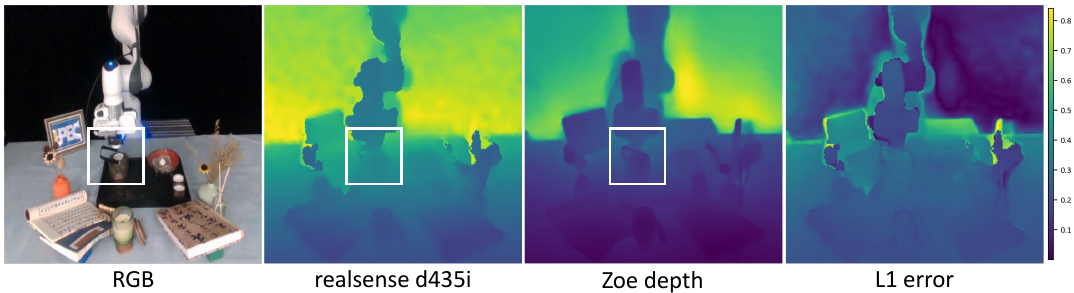}
    \end{center}
    \vspace{-2ex}
    \caption{Depth across sensor, Zoe, and L1 (avg$\le$0.2).}
    \label{fig:depth_noise}
    \vspace{-3ex}
\end{figure}

\subsection{Model Pre-training and Deployment.}
\label{supp:model_pretrain_deploy}
\begin{figure*}[t]
   \begin{center}
      \includegraphics[width=1\textwidth]{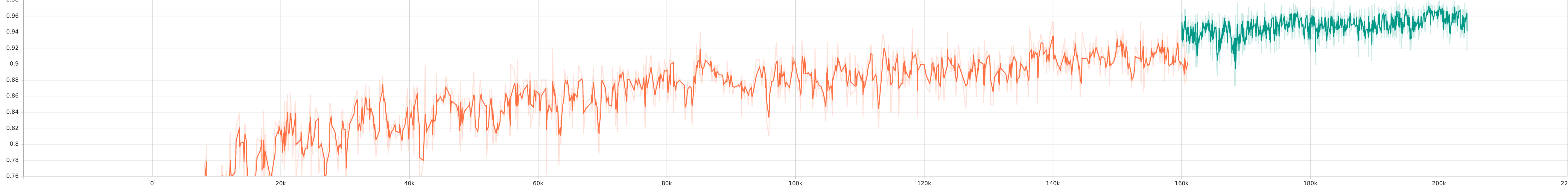} \\
      \vspace{5ex}
      \includegraphics[width=1\textwidth]{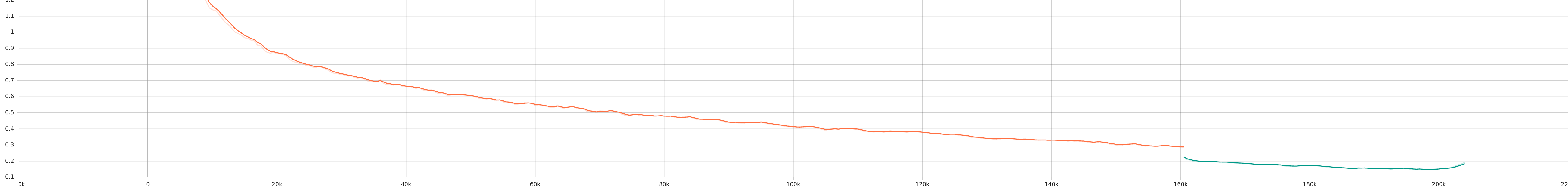}
   \end{center}
   \vspace{-1.5ex}
   \caption{Accuracy (top) and loss (bottom) curves of \ours during training in
   stage 1 (orange) and stage 2 (green).}
   \label{fig:train_accuracy}
\end{figure*}

\noindent
\textbf{Pre-training.} \ours was pre-trained on 64 A100 GPUs for 10 days, with a
batch size of 2048, using AdamW with a learning rate of 2e-5, and a linear scheduler,
with no weight decay and a 0.005 warm-up ratio. The model is initialized from Paligemma2
backbone~\citep{steiner2024paligemma} and fits with 1.1 Million real robot
demonstrations, covering a diverse range of robot embodiments, scenes, and tasks.
Our framework is built entirely on top of the HuggingFace Transformers, with DeepSpeed
integration for memory-efficient and fast-distributed training, utilizing ZeRO
stage 1. The model input consists of a 224$\times$224 image and language
instruction, and the output for each inference is 4 actions with 12 tokens.
\ours training involves two stages. The first stage trains the model on the full
dataset for 160k steps, achieving 90\% accuracy. After removing the DROID dataset
in stage 2, the model is fine-tuned for 40k steps, reaching over 95\%
accuracy. \cref{fig:train_accuracy} shows the training curves, demonstrating a notable
accuracy boost and loss reduction after removing DROID. Notably, data augmentation,
including random crop and color jitter, is critical in enhancing the model's performance,
particularly in the SimplerEnv and WidowX Bridge zero-shot tasks, and real robot
post-training tasks.

\noindent
\textbf{Deployment.} We deploy \ours on the BridgeData V2 WidowX and Franka
robots. The model operates at approximately 20Hz on a single NVIDIA RTX 4090 GPU
with 8.5GB memory consumption for both simulation and real-world evaluations. We
implement ensemble actions with horizons of 1 or 4 as needed, and optimize the
gripper's stick step from 1 to 15 - a critical adjustment for real-world
deployment. For the BridgeV2 WidowX setup, we utilize a WidowX 250 robot
equipped with a statically-mounted Realsense D435 camera providing input images.
The hardware configuration strictly follows Bridge DATA V2~\citep{walke2023bridgedata}
specifications, particularly regarding the D435 camera's positioning relative to
the robotic arm. Since Bridge DATA V2 does not specify exact relative positions
between the robot base and manipulated objects, we meticulously calibrated the robot
arm's placement in the workspace by referencing scenario images from the BridgeDATA
V2 dataset to ensure maximal environmental consistency. For real-world fine-tuning
tasks, we employ a Franka Emika Panda robot with impedance control capabilities~\citep{zhu2023viola}.
During inference, scene perception relies solely on a tripod-mounted Realsense
D435 camera. All fine-tuning datasets are collected via human teleoperation using
a Spacemouse device, sampled at 10Hz.


\subsection{Detailed Evaluation Setup}
\label{supp:eval_setup}

In this section, we provide comprehensive details on the evaluation setups for real-world
experiments using BridgeData V2 WidowX and Franka robots, as discussed in~\cref{sec:zero_shot_robot_control},~\cref{sec:evaluating_spatial_understanding_capability},
and~\cref{sec:new_robot_setup}.

\noindent
\textbf{1. Zero-shot Robot Control Evaluation on WidowX Robot.}

As described in~\ref{sec:zero_shot_robot_control}, we conducted extensive evaluations
of 5 generalist robot manipulation policies across 7 zero-shot tasks, with 11
trials per task on a real-world BridgeV2 WidowX Robot. The specific task settings
are:

\begin{itemize}
   \item \textbf{Close microwave}: The robot must close a toy microwave door positioned
      at various angles (30°, 45°, 60°, and 90°), testing the model's capability
      to manipulate articulated objects in different configurations.

   \item \textbf{Lift red chili pepper}: A basic pick task requiring the robot to
      grasp and lift a red pepper from the sink, designed to evaluate the model's
      object localization accuracy.

   \item \textbf{Put carrot in the plate}: The robot needs to perform a pick-and-place
      task by grasping a carrot from the sink and placing it on a plate,
      assessing both grasping precision and placement accuracy.

   \item \textbf{Put eggplant in the basket}: A complex task requiring the robot
      to identify and pick an eggplant from a sink containing multiple vegetables,
      then place it in a yellow basket. This task evaluates object discrimination
      and spatial awareness.

   \item \textbf{Put purple cup on white plate}: The robot must identify and transfer
      a purple cup to a white plate within a sink containing multiple plates,
      testing color recognition and precise manipulation.

   \item \textbf{Put green cup on pink cloth}: A task involving grasping a green
      cup from the stove and placing it on a pink cloth, serving as a baseline
      for spatial understanding evaluation in~\cref{sec:evaluating_spatial_understanding_capability}.

   \item \textbf{Put purple cup on pink cloth}: Similar to the previous task but
      with a purple cup in the sink, this variation tests the model's ability to
      generalize manipulation skills across different object positions and
      contexts.
\end{itemize}

\noindent
\textbf{2. Adapting to New Robot Setups on Franka Robot.} As described in~\cref{sec:new_robot_setup},
we evaluated the performance of four methods - Diffusion Policy~\citep{chi2024diffusionpolicy},
Octo~\citep{team2024octo}, OpenVLA~\citep{kim2024openvla}, and \ours - across 13
real-world tasks on a Franka Panda Emika robot, with 11 trials per task. While
Diffusion Policy was trained from scratch, Octo, OpenVLA and \ours were fine-tuned
on the specified tasks. The detailed task settings are:

\begin{itemize}
   \item \textbf{Push the teapot handle aside}: The robot must push a teapot handle
      from a perpendicular to a a parallel position relative to the desktop,
      using its gripper tip. This task tests the manipulation of revolute joints
      and includes 50 human demonstrations.

   \item \textbf{Place the white pot on the cutting board}: A pick-and-place task
      requiring the robot to grasp a white bowl from the right side of the table
      and place it on a cutting board, trained with 100 human demonstrations.

   \item \textbf{Put the banana in the basket}: Another pick-and-place task where
      the robot must transfer a banana from the table to a basket. With only 50
      human demonstrations, this task evaluates model performance with limited data.

   \item \textbf{Close the drawer}: Tests the robot's ability to manipulate objects
      with prismatic joints, trained on 100 human demonstrations.

   \item \textbf{Place red/blue/green cube on green car}: An instruction-following
      task where the robot must identify and place a specifically colored cube (red,
      green, or blue) onto a green car. Training included 50 demonstrations for
      each color, totaling 150.

   \item \textbf{Grasp the orange/croissant in the plate}: Another instruction-following
      task requiring the robot to transfer either a croissant (left) or an orange (right)
      to a centrally-placed plate, based on verbal commands. Trained with 200 total
      demonstrations (100 per object).

   \item \textbf{Multi-task of push/place/put/close}: A comprehensive evaluation
      combining the previous manipulation tasks, testing the model's ability to
      handle diverse object interactions and spatial relationships. This extends
      the carrot placement task from~\cref{sec:zero_shot_robot_control} by varying
      target heights and positions.
\end{itemize}

\noindent
\textbf{3. Spatial Understanding Capability Evaluation on Franka and WidowX
Robot.} Following~\cref{sec:evaluating_spatial_understanding_capability}, we
conducted a comprehensive evaluation of spatial understanding capabilities
through 3 zero-shot tasks on the BridgeData V2 WidowX Robot and 1 efficient-finetuning
task on the Franka Robot. The detailed task specifications are:

\begin{itemize}
   \item \textbf{Place plush toy closest to robot on car}: In this efficient fine-tuning
      task, the robot must identify and grasp the nearest plush toy among two
      options and place it on a green car. To rigorously assess spatial understanding,
      we systematically vary the relative positions of the plush toys during
      testing.

   \item \textbf{Put green cup on the pink cloth}: This task suite comprises two
      scenarios testing vertical spatial understanding. In the first scenario, the
      robot grasps a green cup positioned either on a stove or elevated on a yellow
      block. In the second scenario, the cup is placed either at the bottom of a
      sink or elevated on a bowl. This variation in object heights challenges
      the model's ability to adapt its manipulation strategy according to
      spatial configurations.

   \item \textbf{Put carrot in the plate}: Building upon the task in~\cref{sec:zero_shot_robot_control},
      we modify the target plate's position from the sink bottom to an elevated
      position on a pan. This modification specifically tests the model's spatial
      awareness and precision in object placement, requiring accurate perception
      of the plate's pose and appropriate adjustment of the placement trajectory.
\end{itemize}


\subsection{More Detailed Evaluation Results.}
\label{supp:more_eval_results}

\noindent
\textbf{1. SimplerEnv Evaluation.}

\begin{table*}
  [t]
  \centering
  \caption{SimplerEnv evaluation results across different policies on Google
  Robot tasks.}
  \label{tab:simplerenv_google_robot_supp} \resizebox{2\columnwidth}{!}{
  \begin{tabular}{l|lccccccccc}
    \toprule \multirow{5}{*}{\begin{tabular}[l]{@{}l@{}}Google Robot\\ Evaluation Setup\end{tabular}} & \multirow{5}{*}{Policy} & \multicolumn{4}{c}{\multirow{2}{*}{Pick Coke Can}}          & \multirow{2}{*}{Move Near}                                & \multicolumn{3}{c}{\multirow{2}{*}{Open / Close Drawer}} & \multirow{2}{*}{\begin{tabular}[c]{@{}c@{}}\#Overall\end{tabular}} \\
    \\
    \cmidrule{3-11}                                                                                   &                         & \begin{tabular}[c]{@{}c@{}}Horizontal\\ Laying\end{tabular} & \begin{tabular}[c]{@{}c@{}}Vertical\\ Laying\end{tabular} & \begin{tabular}[c]{@{}c@{}}Standing\end{tabular}         & \begin{tabular}[c]{@{}c@{}}Average\end{tabular}                   & \begin{tabular}[c]{@{}c@{}}Average\end{tabular} & \begin{tabular}[c]{@{}c@{}}Open\end{tabular} & \begin{tabular}[c]{@{}c@{}}Close\end{tabular} & \begin{tabular}[c]{@{}c@{}}Average\end{tabular} & \begin{tabular}[c]{@{}c@{}}Average\end{tabular} \\
    \midrule \multirow{10}{*}{\begin{tabular}[l]{@{}l@{}}Variant\\Aggregation\end{tabular}}           & RT-1 (begin)            & 2.2\%                                                       & 1.3\%                                                     & 03.1\%                                                   & 02.2\%                                                            & 04.0\%                                          & 00.5\%                                       & 13.2\%                                        & 06.9\%                                          & 04.2\%                                          \\
                                                                                                      & RT-1 ($15\%$)           & 92.0\%                                                      & 70.4\%                                                    & 81.3\%                                                   & 81.3\%                                                            & 44.6\%                                          & 21.2\%                                       & 32.3\%                                        & 26.7\%                                          & 56.2\%                                          \\
                                                                                                      & RT-1 (converged)        & 96.9\%                                                      & 76.0\%                                                    & 96.4\%                                                   & 89.8\%                                                            & 50.0\%                                          & 27.0\%                                       & 37.6\%                                        & 32.3\%                                          & 63.3\%                                          \\
    \cmidrule{2-11}                                                                                   & TraceVLA                & ---                                                         & ---                                                       & ---                                                      & 60.0\%                                                            & 56.4\%                                          & ---                                          & ---                                           & 31.0\%                                          & 45.0\%                                          \\
                                                                                                      & RT-1-X                  & 56.9\%                                                      & 20.4\%                                                    & 69.8\%                                                   & 49.0\%                                                            & 32.3\%                                          & 06.9\%                                       & 51.9\%                                        & 29.4\%                                          & 39.6\%                                          \\
                                                                                                      & RT-2-X                  & 82.2\%                                                      & 75.4\%                                                    & 89.3\%                                                   & 82.3\%                                                            & 79.2\%                                          & 33.3\%                                       & 37.2\%                                        & 35.3\%                                          & 64.3\%                                          \\
                                                                                                      & Octo-Base               & 0.5\%                                                       & 00.0\%                                                    & 01.3\%                                                   & 00.6\%                                                            & 03.1\%                                          & 00.0\%                                       & 02.1\%                                        & 01.1\%                                          & 01.1\%                                          \\
                                                                                                      & OpenVLA                 & 71.1\%                                                      & 27.1\%                                                    & 65.3\%                                                   & 54.5\%                                                            & 47.7\%                                          & 15.8\%                                       & 19.5\%                                        & 17.7\%                                          & 39.8\%                                          \\
                                                                                                      & RoboVLM (zero-shot)     & 77.8\%                                                      & 48.0\%                                                    & 79.1\%                                                   & 68.3\%                                                            & 56.0\%                                          & 01.6\%                                       & 15.3\%                                        & 08.5\%                                          & 46.3\%                                          \\
                                                                                                      & RoboVLM (fine-tuning)        & 93.8\%                                                      & 49.8\%                                                    & 83.1\%                                                   & 75.6\%                                                            & 60.0\%                                          & 02.6\%                                       & 18.5\%                                        & 10.6\%                                          & 51.3\%                                          \\
    \rowcolor[HTML]{EFEFEF}                                                                           & \ours (zero-shot)       & 93.3\%                                                      & 83.1\%                                                    & 92.0\%                                                   & 89.5\%                                                            & 71.7\%                                          & 23.3\%                                       & 49.2\%                                        & 36.2\%                                          & 68.8\%                                          \\
    \rowcolor[HTML]{EFEFEF}                                                                           & \ours (fine-tuning)          & 93.3\%                                                      & 78.2\%                                                    & 92.4\%                                                   & 88.0\%                                                            & 72.7\%                                          & 28.6\%                                       & 55.0\%                                        & 41.8\%                                          & 70.7\%                                          \\
    \midrule \multirow{13}{*}{\begin{tabular}[l]{@{}l@{}}Visual\\Matching\end{tabular}}               & RT-1 (Begin)            & 5.0\%                                                       & 00.0\%                                                    & 03.0\%                                                   & 02.7\%                                                            & 05.0\%                                          & 00.0\%                                       & 27.8\%                                        & 13.9\%                                          & 06.8\%                                          \\
                                                                                                      & RT-1 ($15\%$)           & 86.0\%                                                      & 79.0\%                                                    & 48.0\%                                                   & 71.0\%                                                            & 35.4\%                                          & 46.3\%                                       & 66.7\%                                        & 56.5\%                                          & 60.2\%                                          \\
                                                                                                      & RT-1 (Converged)        & 96.0\%                                                      & 90.0\%                                                    & 71.0\%                                                   & 85.7\%                                                            & 44.2\%                                          & 60.1\%                                       & 86.1\%                                        & 73.0\%                                          & 74.6\%                                          \\
    \cmidrule{2-11}                                                                                   & HPT                     & ---                                                         & ---                                                       & ---                                                      & 56.0\%                                                            & 60.0\%                                          & ---                                          & ---                                           & 24.0\%                                          & 46.0\%                                          \\
                                                                                                      & TraceVLA                & ---                                                         & ---                                                       & ---                                                      & 28.0\%                                                            & 53.7\%                                          & ---                                          & ---                                           & 57.0\%                                          & 42.0\%                                          \\
                                                                                                      & RT-1-X                  & 82.0\%                                                      & 33.0\%                                                    & 55.0\%                                                   & 56.7\%                                                            & 31.7\%                                          & 29.6\%                                       & 89.1\%                                        & 59.7\%                                          & 53.4\%                                          \\
                                                                                                      & RT-2-X                  & 74.0\%                                                      & 74.0\%                                                    & 88.0\%                                                   & 78.7\%                                                            & 77.9\%                                          & 15.7\%                                       & 34.3\%                                        & 25.0\%                                          & 60.7\%                                          \\
                                                                                                      & Octo-Base               & 21.0\%                                                      & 21.0\%                                                    & 09.0\%                                                   & 17.0\%                                                            & 04.2\%                                          & 00.9\%                                       & 44.4\%                                        & 22.7\%                                          & 16.8\%                                          \\
                                                                                                      & OpenVLA                 & 27.0\%                                                      & 03.0\%                                                    & 19.0\%                                                   & 16.3\%                                                            & 46.2\%                                          & 19.4\%                                       & 51.8\%                                        & 35.6\%                                          & 27.7\%                                          \\
                                                                                                      & RoboVLM (zero-shot)     & 85.0\%                                                      & 43.0\%                                                    & 90.0\%                                                   & 72.7\%                                                            & 66.3\%                                          & 28.7\%                                       & 25.0\%                                        & 26.8\%                                          & 56.3\%                                          \\
                                                                                                      & RoboVLM (fine-tuning)        & 94.0\%                                                      & 47.0\%                                                    & 91.0\%                                                   & 77.3\%                                                            & 61.7\%                                          & 33.3\%                                       & 53.1\%                                        & 43.5\%                                          & 63.4\%                                          \\
    \rowcolor[HTML]{EFEFEF}                                                                           & \ours (zero-shot)       & 70.0\%                                                      & 82.0\%                                                    & 91.0\%                                                   & 81.0\%                                                            & 69.6\%                                          & 49.1\%                                       & 69.4\%                                        & 59.3\%                                          & 71.9\%                                          \\
    \rowcolor[HTML]{EFEFEF}                                                                           & \ours (fine-tuning)          & 85.0\%                                                      & 76.0\%                                                    & 97.0\%                                                   & 86.0\%                                                            & 77.9\%                                          & 50.0\%                                       & 64.8\%                                        & 57.4\%                                          & 75.1\%                                          \\
    \bottomrule
  \end{tabular}
  }
\end{table*}

\cref{tab:simplerenv_google_robot_supp} presents the evaluation results of the
simpler env on the Google robotic task, encompassing tasks such as Coke can manipulation
(horizontal and vertical picking) and drawer operations (opening and closing).
On average, \ours achieves the highest overall visual matching and variant aggregation
performance with a significant margin, demonstrating more generalist manipulation
capabilities.

\noindent
\textbf{2. Zero-shot Robot Control Evaluation on WidowX Robot.}

\begin{table*}
    [h]
    \centering
    \caption{\textbf{Detailed evaluation results on BridgeV2 WidowX zero-shot
    tasks.} SpatialVLA achieves SOTA performance on all 7 real-world BridgeV2 WidowX
    zero-shot tasks, fully demonstrating that SpatialVLA can obtain a large
    amount of pre-training knowledge on heterogeneous embodiment data.}
    \label{tab:widowx_zeroshot} \resizebox{2\columnwidth}{!}{
    \begin{tabular}{l|cccccc}
        \toprule \multirow{2}{*}{Models}             & \multirow{2}{*}{\textbf{Close the microwave}}                               & \multirow{2}{*}{\textbf{Lift the red chili pepper}}                     & \multicolumn{2}{c}{\textbf{Put eggplant in the basket}} & \multicolumn{2}{c}{\textbf{Put purple cup on white plate}} \\
                                                     &                                                                             &                                                                         & Grasp eggplant                                          & Success                                                   & Grasp purple cup & Success          \\
        \cmidrule{1-7} RT-1-X~\citep{o2024open}      & 54.55\%                                                                     & 36.36\%                                                                 & 45.45\%                                                 & 0.00\%                                                    & 18.18\%          & 9.09\%           \\
        Octo-Base~\citep{team2024octo}               & 54.55\%                                                                     & 45.45\%                                                                 & 36.36\%                                                 & 9.09\%                                                    & 0.00\%           & 0.00\%           \\
        RoboVLM~\citep{li2023generalist}             & 72.73\%                                                                     & 0.00\%                                                                  & 27.27\%                                                 & 9.09\%                                                    & 45.45\%          & 27.27\%          \\
        OpenVLA~\citep{kim2024openvla}               & \textbf{100.00\%}                                                           & \textbf{72.72\%}                                                        & 63.63\%                                                 & 63.63\%                                                   & \textbf{63.63\%} & 18.18\%          \\
        \rowcolor[HTML]{EFEFEF} SpatialVLA           & \textbf{100.00\%}                                                           & \textbf{72.72\%}                                                        & \textbf{81.81\%}                                        & \textbf{72.72\%}                                          & 54.54\%          & \textbf{45.45\%} \\
        \midrule \midrule
        \multirow{2}{*}{Models} & \multicolumn{2}{c}{\textbf{Put green cup on the pink cloth (cup on stove)}} & \multicolumn{2}{c}{\textbf{Put purple cup on pink cloth (cup in sink)}} & \multicolumn{2}{c}{\textbf{Put the carrot in the plate}} \\
                                                     & Grasp green cup                                                             & Success                                                                 & Grasp purple cup                                        & Success                                                   & Grasp carrot     & Success          \\
        \cmidrule{1-7} RT-1-X~\citep{o2024open}      & 18.18\%                                                                     & 18.18\%                                                                 & 9.09\%                                                  & 0.00\%                                                    & 9.09\%           & 9.09\%           \\
        Octo-Base~\citep{team2024octo}               & 18.18\%                                                                     & 9.09\%                                                                  & 27.27\%                                                 & 18.18\%                                                   & 27.27\%          & 0.00\%           \\
        RoboVLM~\citep{li2023generalist}             & 27.27\%                                                                     & 18.18\%                                                                 & 18.18\%                                                 & 9.09\%                                                    & 45.45\%          & 27.27\%          \\
        OpenVLA~\citep{kim2024openvla}               & 36.36\%                                                                     & 36.36\%                                                                 & 63.63\%                                                 & 54.54\%                                                   & \textbf{63.63\%} & 18.18\%          \\
        \rowcolor[HTML]{EFEFEF} SpatialVLA           & \textbf{81.81\%}                                                            & \textbf{81.81\%}                                                        & \textbf{90.90\%}                                        & \textbf{90.90\%}                                          & 54.54\%          & \textbf{45.45\%} \\
        \bottomrule
    \end{tabular}}
\end{table*}
\cref{tab:widowx_zeroshot} presents a comprehensive comparison of five
generalist robot manipulation policies' zero-shot performance, including \ours, in
real-world scenarios. For fundamental tasks such as "Close the microwave", we report
the overall task completion success rate. For more sophisticated tasks like "Lift
the red chili pepper", we provide detailed performance metrics by separately evaluating
both the grasping success rate and the complete task execution success rate.
Across all seven real-world evaluation tasks, \ours consistently demonstrated superior
performance, achieving the highest success rates in task completion.

\noindent
\textbf{3. Adapting to New Robot Setups on Franka Robot.}

\begin{table*}
    [t]
    \centering
    \caption{\textbf{Detailed evaluation results on single-task fine-tuning.} On 2 of 4 single-task fine-tuning tasks tested on Franka Emika Panda, SpatialVLA achieves SOTA performance, even surpassing the Diffusion Policy~\citep{chi2024diffusionpolicy} trained from scratch.} 
    \label{tab:single_task_finetune} 
    \resizebox{2\columnwidth}{!}{
    \begin{tabular}{l|cccccc}
        \toprule \multirow{2}{*}{Models}                               & \textbf{Push the Teapot Handle Aside} & \multicolumn{2}{c}{\textbf{Place the White Pot on the Cutting Board}} & \multicolumn{2}{c}{\textbf{Put the Banana in the Basket}} & \textbf{Close the Drawer} \\
                                                                       & Success                               & Grasp pot                                                             & Success                                                   & Grasp banana             & Success         & Success         \\
        \cmidrule{1-7} Diffusion Policy~\citep{chi2024diffusionpolicy} & 72.72\%                                 & \textbf{81.81\%}                                                        & \textbf{72.72\%}                                            & 72.72\%                    & 72.72\%           & \textbf{100.00\%} \\
        Octo~\citep{team2024octo}                                      & \textbf{100.00\%}                       & 72.72\%                                                                 & 63.63\%                                                     & 0.00\%                     & 0.00\%            & \textbf{100.00\%} \\
        OpenVLA~\citep{kim2024openvla}                                 & \textbf{100.00\%}                       & 63.63\%                                                                 & 54.54\%                                                     & 81.81\%                    & 54.54\%           & 81.81\%           \\
        \rowcolor[HTML]{EFEFEF} SpatialVLA                             & 81.81\%                                 & 72.72\%                                                                 & \textbf{72.72\%}                                            & \textbf{100.00\%}          & \textbf{100.00\%} & 72.72\%           \\
        \bottomrule
    \end{tabular}
    }
\end{table*}

\begin{table*}
    [t]
    \centering
    \caption{\textbf{Detailed evaluation results on multi-task fine-tuning.} On 3
    of 4 multi-task fine-tuning tasks tested on Franka Emika Panda, SpatialVLA
    achieves SOTA performance, improving its powerful multi-task learning capabilities.}
    \label{tab:multi_task_finetuning} \resizebox{2\columnwidth}{!}{
    \begin{tabular}{l|cccccc}
        \toprule \multirow{2}{*}{Models}                               & \textbf{Push the Teapot Handle Aside} & \multicolumn{2}{c}{\textbf{Place the White Pot on the Cutting Board}} & \multicolumn{2}{c}{\textbf{Put the Banana in the Basket}} & \textbf{Close the Drawer} \\
                                                                       & Success                               & Grasp pot                                                             & Success                                                   & Grasp banana             & Success          & Success          \\
        \cmidrule{1-7} Diffusion Policy~\citep{chi2024diffusionpolicy} & 18.18\%                               & 36.36\%                                                               & 27.27\%                                                   & 27.27\%                  & 27.27\%          & 27.27\%          \\
        Octo~\citep{team2024octo}                                      & 27.27\%                               & 36.36\%                                                               & 36.36\%                                                   & 45.45\%                  & 36.36\%          & 27.27\%          \\
        OpenVLA~\citep{kim2024openvla}                                 & 54.54\%                               & \textbf{72.72\%}                                                      & \textbf{54.54\%}                                          & \textbf{81.81\%}         & 63.63\%          & 36.36\%          \\
        \rowcolor[HTML]{EFEFEF} SpatialVLA                             & \textbf{63.63\%}                      & 54.54\%                                                               & 36.36\%                                                   & \textbf{81.81\%}         & \textbf{81.81\%} & \textbf{45.45\%} \\
        \bottomrule
    \end{tabular}
    }
\end{table*}

\begin{table*}
    [t]
    \centering
    \caption{\textbf{Detailed evaluation results on instruction following fine-tuning
    tasks.} On 3 of 5 instruction following fine-tuning tasks tested on Franka Emika
    Panda, SpatialVLA achieves SOTA performance, improving its powerful instruction
    following capabilities.}
    \label{tab:instruction_following} \resizebox{2\columnwidth}{!}{
    \begin{tabular}{l|cccccccccc}
        \toprule \multirow{2}{*}{Models}                                & \multicolumn{2}{c}{\textbf{Put Red Cube on Green Car}} & \multicolumn{2}{c}{\textbf{Put Green Cube on Green Car}} & \multicolumn{2}{c}{\textbf{Put Blue Cube on Green Car}} & \multicolumn{2}{c}{\textbf{Grasp the Orange in the Plate}} & \multicolumn{2}{c}{\textbf{Grasp the Croissant in the Plate}} \\
                                                                        & Grasp red cube                                         & Success                                                  & Grasp green cube                                        & Success                                                    & Grasp blue cube                                              & Success          & Grasp orange     & Success          & Grasp croissant  & Success          \\
        \cmidrule{1-11} Diffusion Policy~\citep{chi2024diffusionpolicy} & 18.18\%                                                & 9.09\%                                                   & 0.00\%                                                  & 0.00\%                                                     & 0.00\%                                                       & 0.00\%           & 27.27\%          & 18.18\%          & 27.27\%          & 9.09\%           \\
        Octo~\citep{team2024octo}                                       & \textbf{63.63\%}                                       & \textbf{63.63\%}                                         & 0.00\%                                                  & 0.00\%                                                     & 45.45\%                                                      & 45.45\%          & \textbf{72.72\%} & \textbf{63.63\%} & 27.27\%          & 18.18\%          \\
        OpenVLA~\citep{kim2024openvla}                                  & 54.54\%                                                & 54.54\%                                                  & \textbf{54.54\%}                                        & 36.36\%                                                    & 45.45\%                                                      & 27.27\%          & 36.36\%          & 27.27\%          & 45.45\%          & 36.36\%          \\
        \rowcolor[HTML]{EFEFEF} SpatialVLA                              & \textbf{63.63\%}                                       & 54.54\%                                                  & 45.45\%                                                 & \textbf{45.45\%}                                           & \textbf{72.72\%}                                             & \textbf{63.63\%} & 63.63\%          & 54.54\%          & \textbf{54.54\%} & \textbf{54.54\%} \\
        \bottomrule
    \end{tabular}
    }
\end{table*}

\cref{tab:single_task_finetune}~\cref{tab:multi_task_finetuning} and~\cref{tab:instruction_following}
present comprehensive evaluation results comparing the performance of four state-of-the-art
approaches: Diffusion Policy~\citep{chi2024diffusionpolicy}, Octo~\citep{team2024octo},
OpenVLA~\citep{kim2024openvla}, and our proposed \ours on the Franka Panda Emika
robot. For systematic analysis, we categorize the evaluation results into three
distinct scenarios: single-task fine-tuning, multi-task fine-tuning, and instruction-following
fine-tuning tasks.

\cref{tab:single_task_finetune} details the model performance in single-task fine-tuning
scenarios. Our proposed \ours achieved state-of-the-art performance on 2
out of 4 tasks, demonstrating its strong capability in individual task adaptation.
While it performed marginally behind the Diffusion Policy (trained from scratch)
on the "Close the Drawer" task, SpatialVLA showed superior performance compared to
Diffusion Policy on the challenging "Push the Teapot Handle Aside" task, though
slightly trailing behind the fine-tuned versions of Octo and OpenVLA. These results
highlight the competitive performance of SpatialVLA in single-task scenarios while
revealing areas for potential improvement.

\cref{tab:multi_task_finetuning} showcases the comparative analysis of all four models
in multi-task fine-tuning settings. Notably, in these more challenging scenarios
requiring generalization across multiple tasks, the Diffusion Policy trained
from scratch exhibited a significant performance degradation. This observation aligns
with findings reported in~\citep{kim2024openvla}, which indicate that generalist robot
operation policies consistently outperform scratch-trained Diffusion Policy in multi-task
evaluations. SpatialVLA demonstrated remarkable performance, achieving the highest
success rate in 3 out of 4 tasks, with only a marginal performance gap compared
to OpenVLA in the "Place the White Pot on the Cutting Board" task. These results
underscore SpatialVLA's robust generalization capabilities across diverse manipulation
scenarios.

\cref{tab:instruction_following} presents a detailed evaluation of the models'
performance in instruction-following fine-tuning tasks, which require both physical
manipulation skills and language understanding capabilities. In this challenging
domain, SpatialVLA exhibited exceptional performance, achieving the highest
success rate in 3 out of 5 tasks. This stands in stark contrast to the Diffusion
Policy trained from scratch, which showed significant performance deterioration
and demonstrated an inability to execute tasks based on language
instructions effectively. These results highlight SpatialVLA's superior ability to integrate
language understanding with physical manipulation, a crucial capability for real-world
robotic applications.

\noindent
\textbf{4. Spatial Understanding Capability Evaluation on Franka and WidowX
Robot}

\begin{table*}
    [t]
    \centering
    \caption{\textbf{Detailed evaluation results on spatial understanding
    capabilities on fine-tuning and zero-shot tasks.} SpatialVLA achieved SOTA performance
    in the first fine-tuning task and the three subsequent zero-shot tasks,
    fully demonstrating the model's powerful spatial understanding capabilities
    with the help of egocentric 3D Spatial Representations and Adaptive Spatial
    Action Grids.}
    \label{tab:sptaial_understanding} \resizebox{2\columnwidth}{!}{
    \begin{tabular}{l|cccccccc}
        \toprule \multirow{2}{*}{Models}                               & \multicolumn{2}{c}{\textbf{Place Plush Toy Closest to Robot on Car}} & \multicolumn{2}{c}{\textbf{Put Green Cup on Pink Cloth (Stove)}} & \multicolumn{2}{c}{\textbf{Put Green Cup on Pink Cloth (Sink)}} & \multicolumn{2}{c}{\textbf{Put Carrot in Plate}} \\
                                                                       & Grasp plush toy                                                      & Success                                                          & Grasp green cup                                                 & Success                                         & Grasp green cup  & Success          & Grasp carrot     & Success          \\
        \cmidrule{1-9} Diffusion Policy~\citep{chi2024diffusionpolicy} & 45.45\%                                                              & 27.27\%                                                          & ---                                                               & ---                                               & ---                & ---                & ---                & ---                \\
        RT-1-X~\citep{o2024open}                                       & ---                                                                    & ---                                                                & 18.18\%                                                         & 9.09\%                                          & 18.18\%          & 0.00\%           & 9.09\%           & 0.00\%           \\
        Octo-Base~\citep{team2024octo}                                 & 63.63\%                                                              & 27.27\%                                                          & 9.09\%                                                          & 9.09\%                                          & 18.18\%          & 18.18\%          & 18.18\%          & 9.09\%           \\
        RoboVLM~\citep{li2023generalist}                               & ---                                                                    & ---                                                                & 9.09\%                                                          & 0.00\%                                          & 9.09\%           & 9.09\%           & 9.09\%           & 0.00\%           \\
        OpenVLA~\citep{kim2024openvla}                                 & 45.45\%                                                              & 45.45\%                                                          & 36.36\%                                                         & 27.27\%                                         & 54.54\%          & 45.45\%          & 54.54\%          & 54.54\%          \\
        \rowcolor[HTML]{EFEFEF} SpatialVLA                             & \textbf{72.72\%}                                                     & \textbf{63.63\%}                                                 & \textbf{81.81\%}                                                & \textbf{72.72\%}                                & \textbf{81.81\%} & \textbf{81.81\%} & \textbf{72.72\%} & \textbf{63.63\%} \\
        \bottomrule
    \end{tabular}}
\end{table*}

\cref{tab:sptaial_understanding} presents a comprehensive evaluation of spatial understanding
capabilities across different models and task scenarios. The evaluation consists
of one fine-tuning task, "Place Plush Toy Closest to Robot on Car", and three
zero-shot tasks designed to test spatial reasoning abilities. In the fine-tuning
scenario, we compared SpatialVLA against several state-of-the-art baselines: the
Diffusion Policy trained from scratch, the fine-tuned Octo-Base model, and the fine-tuned
OpenVLA system. For zero-shot evaluation, we benchmarked SpatialVLA against pre-trained
versions of RT-1-X, Octo-Base, RoboVLM, and OpenVLA. The results demonstrate that
SpatialVLA consistently outperforms all baseline methods across every spatial understanding
task. This superior performance can be attributed to two key architectural
innovations: (1) the incorporation of egocentric 3D Spatial Representations,
which enables better understanding of spatial relationships in the robot's
environment, and (2) the implementation of Adaptive Spatial Action Grids, which
allows for more precise and context-aware action generation. These components work
in concert to enable SpatialVLA to achieve the highest success rates across all evaluated
tasks that demand sophisticated spatial reasoning capabilities, from basic
object manipulation to complex spatial relationship understanding.

\subsection{Visulizatiob of Dataset Statistic.}
\label{supp:stat} 
\begin{figure}[htbp]
   \begin{center}
      \includegraphics[width=0.98\linewidth]{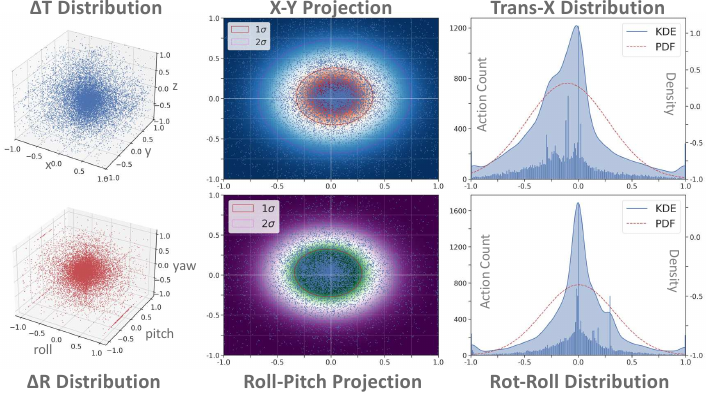}
   \end{center}
   \caption{SpatialVLA Training Dataset Statistic.}
   \label{fig:dataset_statistic}
\end{figure}

~\cref{fig:dataset_statistic} shows the normalized action visualization
of \ours pre-training dataset, where the translation and rotation scatter plots
form an ellipsoidal shape, clustering around the center. Furthermore, the X-Y and
roll-pitch projection plots reveal that almost all data samples cluster within the
2-sigma action space, radiating outward. This suggests that uniform encoding of the
entire [-1,1] range is unnecessary. Instead, adaptive division of the action
space based on sampling probability can efficiently encode fine-grained
operations while reducing encoding costs. Additionally, as discussed in the
Limitations\cref{sec:conclusion}, while Gaussian distributions may not be the
most accurate, we adopt them in this paper for the sake of generality and simplicity,
rather than using kernel density estimations (KDE).

\end{document}